\documentclass[letterpaper, 10 pt, conference]{ieeeconf}  

\IEEEoverridecommandlockouts                              

\usepackage{graphicx}
\usepackage{epstopdf}
\usepackage{amsmath}
\usepackage{amssymb}
\usepackage{tikz} 
\usepackage[normalem]{ulem}
\usepackage{array}
\usepackage{booktabs}

\usepackage{amsthm}
\usepackage[caption=false,font=footnotesize]{subfig}
\let\labelindent\relax
\usepackage[inline]{enumitem}
\usepackage{multirow}
\usepackage{lipsum}
\usepackage{color}
\usepackage[hidelinks]{hyperref}

\newcommand*\rot{\rotatebox{90}}
\newcolumntype{L}[1]{>{\raggedright\let\newline\\\arraybackslash\hspace{0pt}}m{#1}}
\newcolumntype{C}[1]{>{\centering\let\newline\\\arraybackslash\hspace{0pt}}m{#1}}
\newcolumntype{R}[1]{>{\raggedleft\let\newline\\\arraybackslash\hspace{0pt}}m{#1}}

\title{\LARGE \bf
From Motions to Emotions: Can the Fundamental Emotions be Expressed in a Robot Swarm?
}

\author{Mar\'ia Santos and Magnus Egerstedt
\thanks{This work was supported by ``la Caixa'' Banking Foundation under Grant LCF/BQ/AA16/11580039.}
\thanks{The authors are with the School of Electrical and Computer Engineering, Georgia Institute of Technology, Atlanta, Georgia, USA 
        {\tt\small \{\href{mailto:maria.santos@gatech.edu}{\tt\small maria.santos},\href{mailto:magnus@gatech.edu}{\tt\small magnus}\}@gatech.edu}}
}

\begin{document}

\maketitle
\thispagestyle{empty}
\pagestyle{empty}

\begin{abstract}
This paper explores the expressive capabilities of a swarm of miniature mobile robots within the context of inter-robot interactions and their mapping to the so-called fundamental emotions. In particular, we investigate how motion and shape descriptors that are psychologically associated with different emotions can be incorporated into different swarm behaviors for the purpose of artistic expositions. Based on these characterizations from social psychology, a set of swarm behaviors is created, where each behavior corresponds to a fundamental emotion. The effectiveness of these behaviors is evaluated in a survey in which the participants are asked to associate different swarm behaviors with the fundamental emotions. The results of the survey show that most of the research participants assigned to each video the emotion intended to be portrayed by design. These results confirm that abstract descriptors associated with the different fundamental emotions in social psychology provide useful motion characterizations that can be effectively transformed into expressive behaviors for a swarm of simple ground mobile robots.
\end{abstract}

\section{Introduction}
\label{sec:introduction}
Robots have progressively migrated from purely industrial environments
to more social settings where they interact with humans in quotidian activities such as education \cite{Brown2013}, companionship \cite{Belpaeme2013,Hoffman2013}, or health care and therapy \cite{Cabibihan2013,Kozima2009}. In these scenarios, on top of performing tasks related to the specific application, there may be a need for the robots to effectively interact with people in an entertaining, engaging, or anthropomorphic manner \cite{Breazeal2003}.

The need for enticing interactions between social robots and humans becomes especially pronounced in artistic applications. Robots have been progressively intertwined with different forms of artistic expression, where they are used, among others, to interactively create music \cite{Hoffman2010}, dance \cite{Bi2018,LaViers2014,Nakazawa2002,Shinozaki2008}, act in plays \cite{Lee2014,Perkowski2005,Sunardi2018}, support performances \cite{Ackerman2014}, or be the object of art exhibits by themselves \cite{Dean2008,Dunstan2016,Vlachos2018}. As in the traditional expressions of these performing arts, where human artists instill expressive and emotional content \cite{Camurri2004,Juslin2005}, robots are required to convey artistic expression and emotion through their actions. 

While expressive interactions have been extensively studied in the context of performing arts, the focus has been primarily on anthropomorphic robots, especially humanoids \cite{Lee2014,Or2009,Perkowski2013}. However, for faceless robots or robots with limited degrees of freedom for which mimicking human movement is not an option, creating expressive behaviors can pose increased difficulty \cite{Bretan2015,Hoffman2008,Schoellig2014}. We are interested in exploring the expressive capabilities of a swarm of miniature mobile robots, for which the study of expressive interactions is sparse \cite{Dietz2017,Levillain2018,StOnge2019}. This can be contrasted with more anthropomorphic robots, for which there is already a preconceived understanding of emotive expressiveness. This choice is driven in part by the increased prevalence of multi-robot applications and the envisioned, resulting large-scale human-robot teams \cite{Goodrich2007HRIsurvey,Kolling2016,Sheridan2016}; and in part by the expressive possibilities of the swarm as a collective in contrast to the robots as individuals. While using teams of mobile robots to create artistic effects in performances is not something new \cite{Ackerman2014,Alonso-Mora2014}, our aim is to provide a framework to use these types of robotic teams in performances without the need for a choreographer to specify the parameters of the robots' movements, as in \cite{Schoellig2014}. 

\begin{table*}[t]
\caption{Movement and shape attributes associated with different emotions and emotion valences.}
\label{tab:motion_shape_emotion}
 \centering
 \renewcommand*{\arraystretch}{1.2}
 \begin{tabular}{clccc}
 \hline
 
  & &  Shape Features & Size Features & Movement Features\\
  \hline
  \multirow{ 5}{*}{Emotion} & Happiness & roundness, curvilinearity \cite{Collier1996} & big \cite{DeRooij2013} & smoothness \cite{Lee2007}\\
  & Surprise &  roundness \cite{Collier1996} & very big \cite{DeRooij2013} &\\
  & Sadness & roundness \cite{Collier1996} & small \cite{DeRooij2013} & small, slow \cite{Pollick2001,Rime1985} \\
  & Anger & &  & large, fast, angular \cite{Pollick2001}\\
  & Fear & downward pointing triangles \cite{Aronoff2006} & & small, slow \cite{Pollick2001,Rime1985} \\
  \hline
  \multirow{2}{*}{Valence} & Positive & roundness \cite{Aronoff2006,Collier1996} & & rounded movement trace \cite{Aronoff2006,Collier1996}\\
  & Negative & angularity \cite{Aronoff2006,Collier1996} & & angular movement trace \cite{Aronoff2006,Collier1996}\\
  \hline
 \end{tabular}
\end{table*}

Social psychology has extensively studied which motion and shape descriptors are associated with different fundamental emotions, e.g. \cite{Collier1996,Ekman1993,Lee2007,Pollick2001,Rime1985}. In this paper, we study how such attributes can be incorporated into the movements of a swarm of mobile robots to represent  emotions. In particular, a series of swarm behaviors associated with the so-called fundamental emotions are designed and evaluated in a user study in order to determine if a human can identify the different fundamental emotions by observing the swarm aggregate behavior and movement of the individual robots.

The paper is organized as follows: In Section \ref{sec:psychology}, we outline the motion and shape characteristics psychologically linked to the different fundamental emotions. The behaviors included in the user study, implemented on the swarm according to the features described in the social psychology literature, are characterized in Section \ref{sec:swarm_behavior_design}. The procedure and results of the study conducted with human subjects are presented in Section \ref{sec:user_study}, along with the discussion. An implementation of the proposed swarm behaviors on a real robotic platform is presented in Section \ref{sec:robotic_implementation}. Section \ref{sec:conclusions} concludes the paper. 

\section{Emotionally Expressive Movement}
\label{sec:psychology}
For robotic swarms to participate in artistic expositions and effectively convey emotional content, the swarm's behavior when depicting a particular emotion should be recognizable by the audience, thus producing the effect intended by the artist. However, the lack of anthropomorphism in a robotic swarm can pose a challenge when creating expressive motions for human spectators. In this section, we present a summary of motion and shape features that have been linked to different emotions in the social psychology literature, which will serve as inspiration to create expressive behaviors for swarms of mobile robots. 

In this study, we focus on the so-called \textit{fundamental emotions} \cite{Ekman1993,Izard2009}---i.e. happiness, sadness, anger, fear, surprise and disgust---to produce a tractable set of emotion behaviors to be executed by the robotic swarm. An emotion is considered \textit{fundamental} or \textit{basic} if it is inherent to human mentality and adaptive behavior, and remains recognizable across cultures \cite{Izard1977}. In addition, fundamental emotions provide a basis for a wider range of human emotions, which appear at the intersection of the basic emotions with varying intensities \cite{Plutchik2001}.

\begin{figure}[t!]
 \centering
 \includegraphics[width=0.5\linewidth]{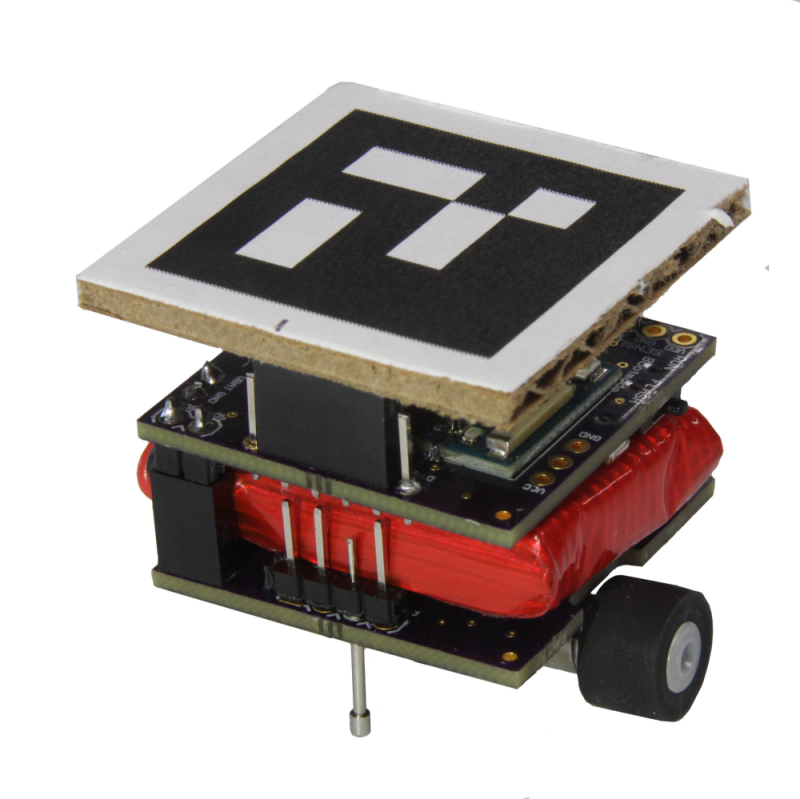}
 \caption{The GRITSBot, a 3cmx3cm miniature mobile differential drive robot. The robotic swarm considered in this study is composed of 15 GRITSBots. The top view of these robots is used in the simulations shown to the study participants when evaluating the different swarm behaviors.}
 \label{fig:gritsbot}
\end{figure}

The robotic system considered for this study is a swarm of miniature differential-drive robots, the GRITSBots \cite{Pickem15}. As shown in Fig. \ref{fig:gritsbot}, the GRITSBots are faceless robots that do not possess any anthropomorphic features. While Laban Movement Analysis \cite{Laban1947} has been used in robotic systems to convey emotional content through acceleration patterns \cite{Barakova2010,Knight2014,Lourens2010,Masuda2010}, when considering large robot swarms, the individual robots may be limited in size and actuation capabilities, thereby restricting their ability to use acceleration as their expressive means. For this reason, along with the characteristic non-anthropomorphism of a swarm and the possibilities of its collective behavior, we draw inspiration from abstract shape and motion descriptors associated with different fundamental emotions \cite{DeRooij2013} to create different swarm behaviors. 

Table \ref{tab:motion_shape_emotion} presents a summary of the shape, movement and size attributes of abstract objects associated with some fundamental emotions and emotion valences. Among these characterizations, those related to shape and size represent the impact of the form of an object on its emotion attribution. In particular, angular shape contours are typically associated with emotions with a negative valence and high arousal\footnote{In this context, the term \emph{valence} designates the intrinsic attractiveness (positive valence) or aversiveness (negative valence) of an event, object, or situation \cite{Frijda1986}. The valence of an emotion thus characterizes its positive or negative connotation. Among the fundamental emotions, happiness and surprise have positive valence, while the remaining four---sadness, fear, disgust and anger---are classified under negative valence \cite{Russell1980}. On the other hand, the term \emph{arousal} refers the activation or deactivation associated with an emotion.}---i.e. anger, fear and disgust---while round shape contours are linked to positive emotions (happiness and surprise) or emotions with very low activation levels (sadness) \cite{Aronoff2006,Collier1996}. The size of a particular object also affects its emotional perception, with bigger objects being typically associated with larger emotion arousal (e.g. surprise) and smaller sizes with emotions with low activation \cite{DeRooij2013}. Table \ref{tab:motion_shape_emotion} also presents how the features of different movement patterns are related to perceived emotions \cite{Pollick2001}. Analogously to shape contours, smoothness of movement is related to the pleasantness of the motion, thus evoking emotions with positive valence \cite{Lee2007}, while an angular movement trace---interpreted as the trajectory taken by the robot over time---is linked to negative emotions \cite{Pollick2001}. Speed of movement also influences the emotion attribution, with higher peak velocities being identified with angry states \cite{Pollick2001} and slower movements that integrate into smaller trajectories over time being connected to fearful and sad emotional states \cite{Pollick2001,Rime1985}.

While the summary of features related to emotions in Table \ref{tab:motion_shape_emotion} provides a good starting point for generating swarm behaviors for most fundamental emotions, literature on motion characterizations of \textit{disgust} is scarce. In order to get some intuition about which traits the swarm behavior should portray when embodying this emotion, we direct our attention towards characterizations associated with emotion valence. The shape and motion characterizations of positive and negative emotion valences in the lower part of Table \ref{tab:motion_shape_emotion} serve as a basis to design the swarm behavior associated with disgust.

\begin{figure}[t!]
 \centering
 \includegraphics[width=\linewidth]{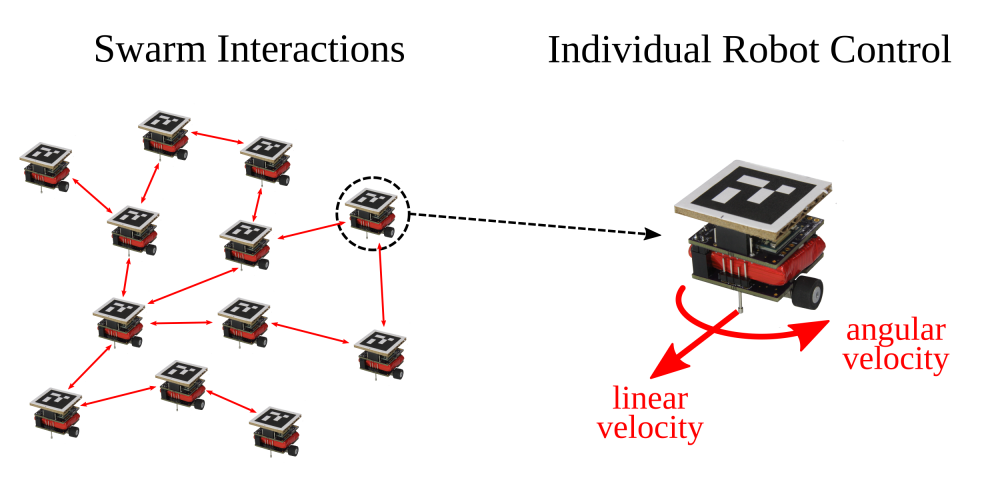}
 \caption{The behavior of a robotic swarm depends on which interactions are considered between the robots, which information is exchanged through those interactions, and how each robot acts on such information. Different interaction schemes and control laws produce distinctly different swarm behaviors.}
 \label{fig:swarmVsIndividual}
\end{figure}

The behavior of a robotic swarm depends on how the interactions are established between members of the swarm and what control commands are executed by the individuals based on the information exchanged in those interactions, as illustrated in Fig. \ref{fig:swarmVsIndividual}. While the GRITSBots as individuals cannot change their shape, the collective behavior of the swarm may embody the \textit{shape} and \textit{size} attributes included in Table \ref{tab:motion_shape_emotion}. On the other hand, the \textit{movement} features in Table \ref{tab:motion_shape_emotion} can be depicted through the movement trace that each individual robot executes as it progresses towards the collective shape. In the next section, we describe how all these attributes are implemented in the controller of the robots to produce the behaviors that embody the different fundamental emotions. 

\section{Swarm Behavior Design}
\label{sec:swarm_behavior_design}
For our swarm of robots to be expressive, we need to decide which interactions a robot should establish with the robots in its vicinity and its environment, and which control law the robot should execute with the information obtained through those interactions to produce an appropriate swarm behavior. In this paper, we draw inspiration from standard algorithms for multi-robot teams, namely cyclic pursuit \cite{Justh2003,Marshall2004,Ramirez2009} and coverage control \cite{Cortes04,DiazMercado2015}, to design the interactions and the control laws for the swarm. This section describes how the shape and movement features described in Section \ref{sec:psychology} are incorporated into the control laws of a swarm of 15 GRITSBots in order to create expressive behaviors. 

\begin{figure*}[t!]
	\centering%
	\subfloat[\label{subfig:happinessf1}$t =0$ s]{\includegraphics[width=0.33\linewidth]{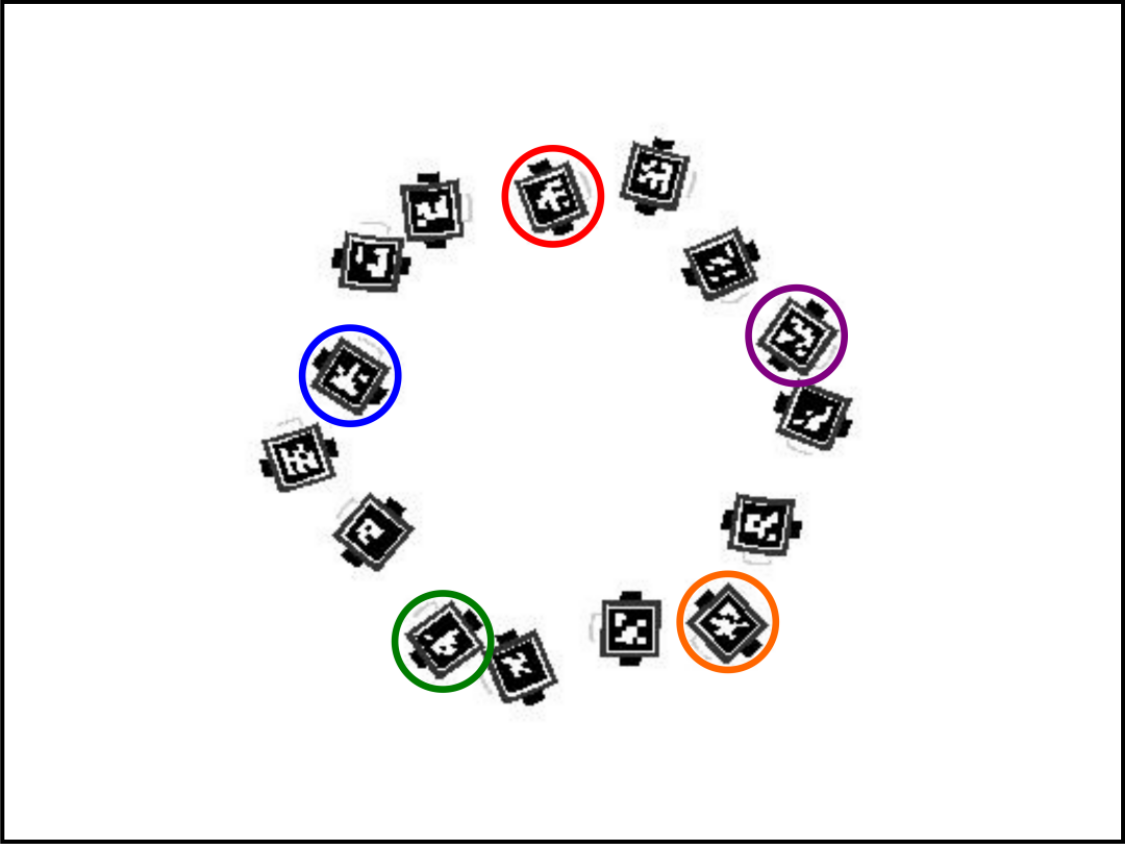}}%
	\subfloat[\label{subfig:happinessf2}$t =1$ s]{\includegraphics[width=0.33\linewidth]{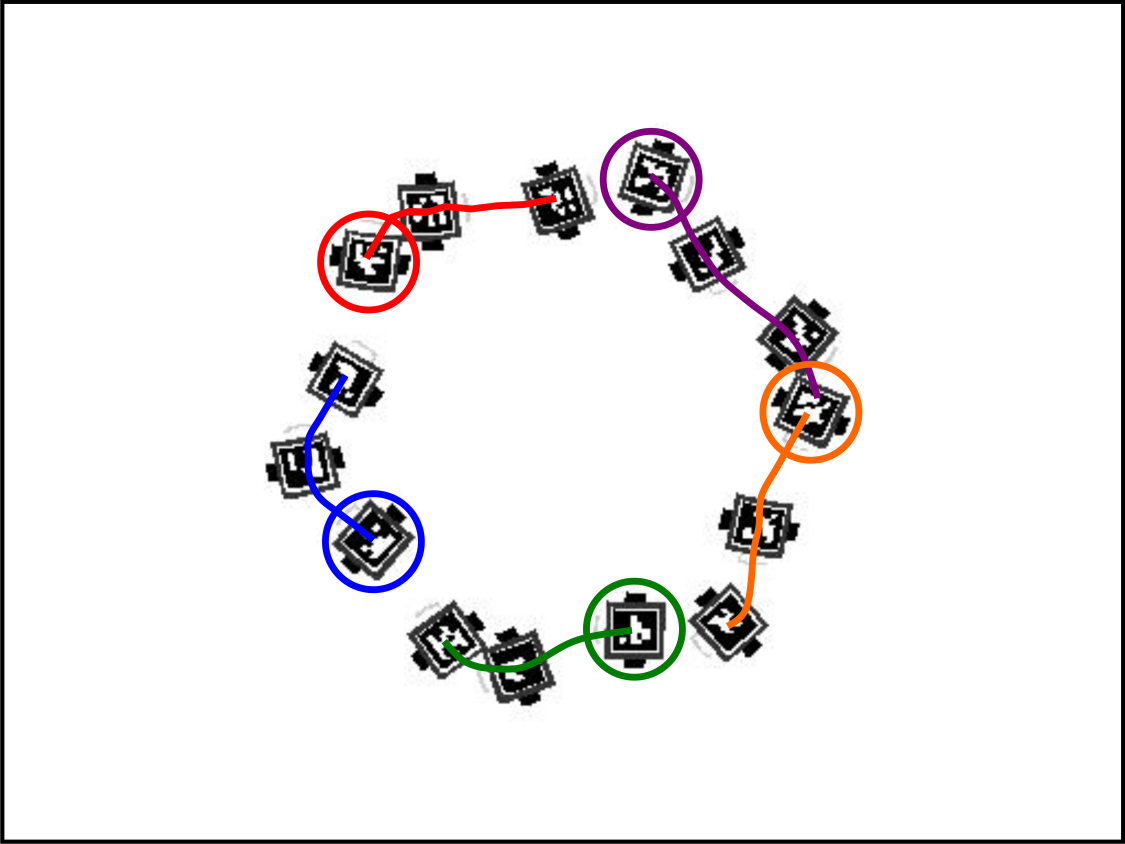}}%
	\subfloat[\label{subfig:happinessf3}$t =4$ s]{\includegraphics[width=0.33\linewidth]{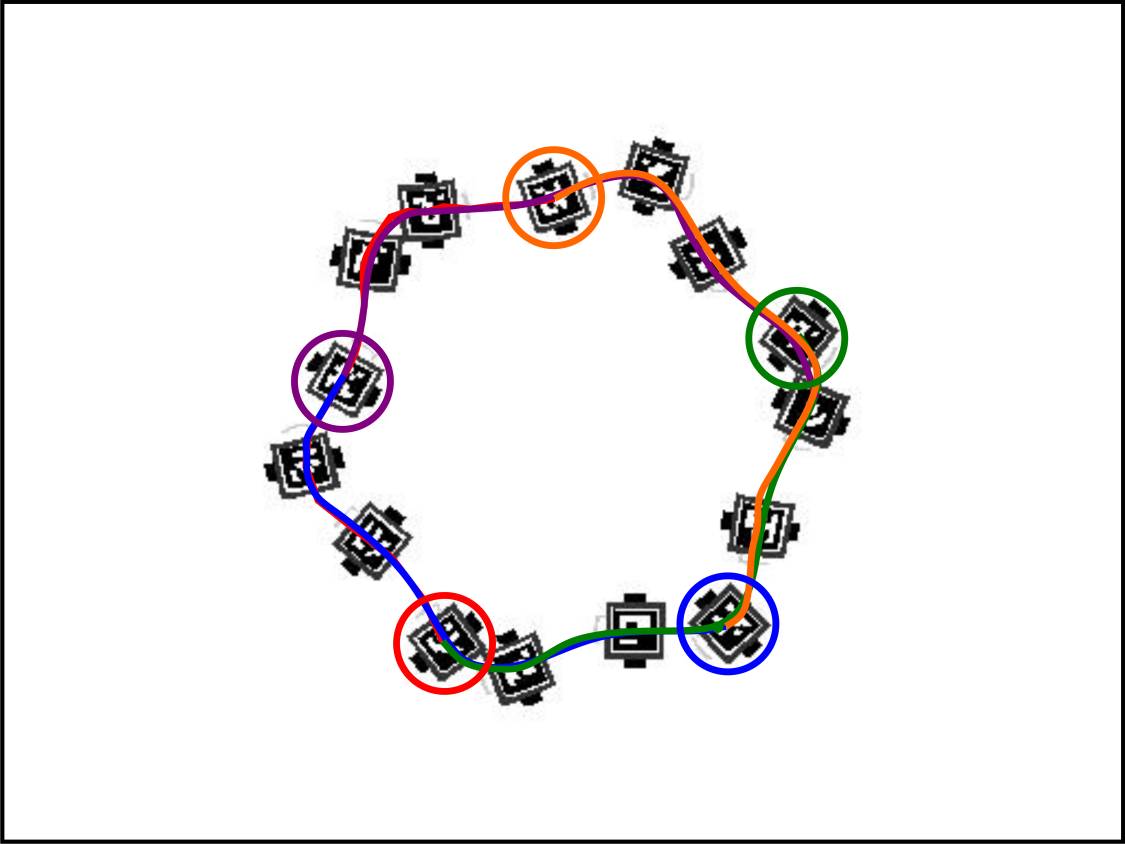}}%
	\caption{Sequence of snapshots of the \textit{happiness} behavior. Each robot follows a point that travels along a circular sinusoid, visually producing a circular shape with small ripples. The trajectories of five robots have been plotted using solid lines. See the full video at \url{https://youtu.be/q_FenI1DdRY}.}
	\label{fig:happinessFrames}
\end{figure*}

\begin{figure*}[t!]
	\centering%
	\subfloat[\label{subfig:surprisef1}$t =0$ s]{\includegraphics[width=0.33\linewidth]{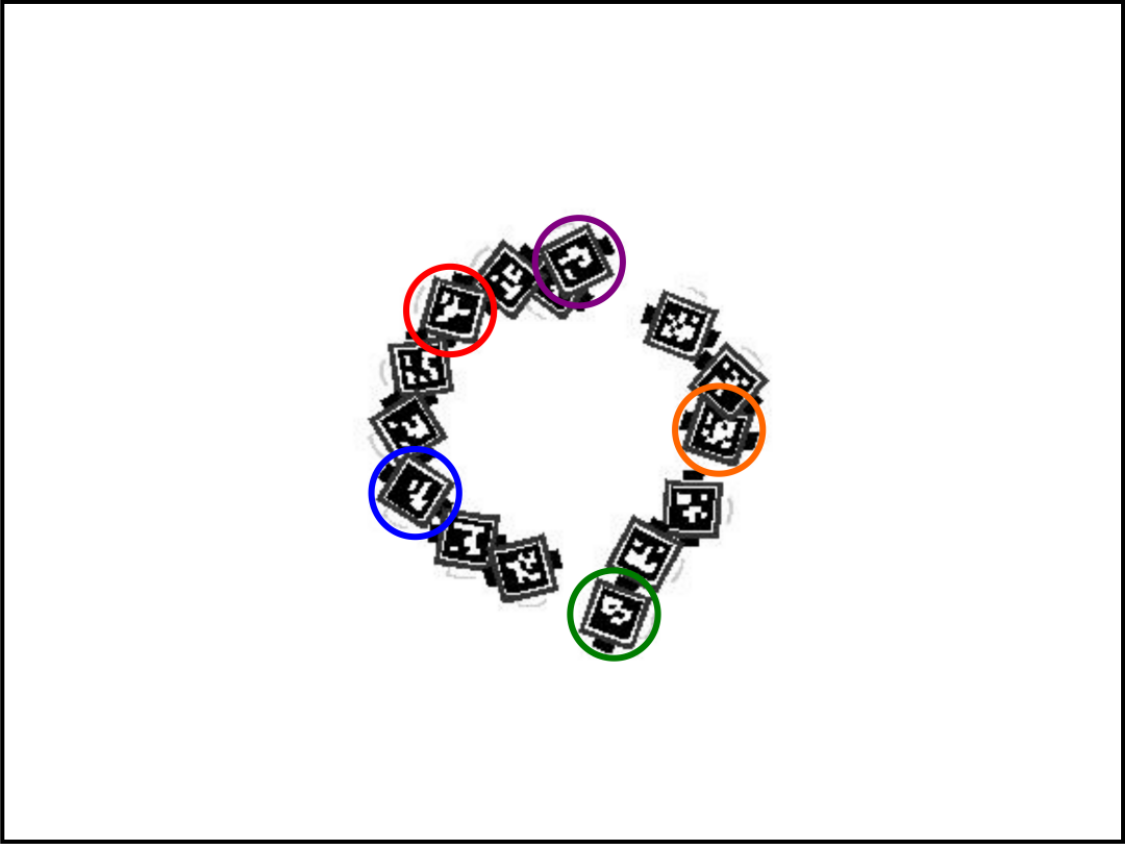}}%
	\subfloat[\label{subfig:surprisef2}$t =1$ s]{\includegraphics[width=0.33\linewidth]{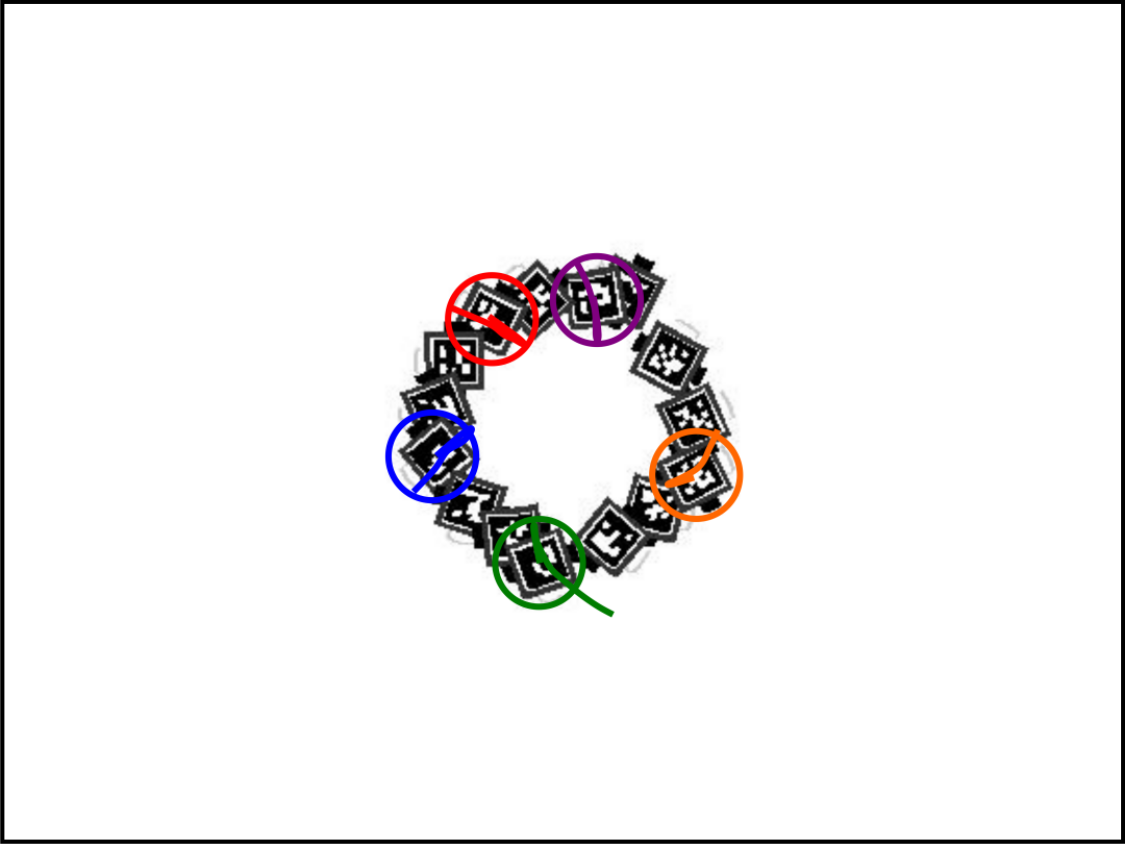}}%
	\subfloat[\label{subfig:surprisef3}$t =4$ s]{\includegraphics[width=0.33\linewidth]{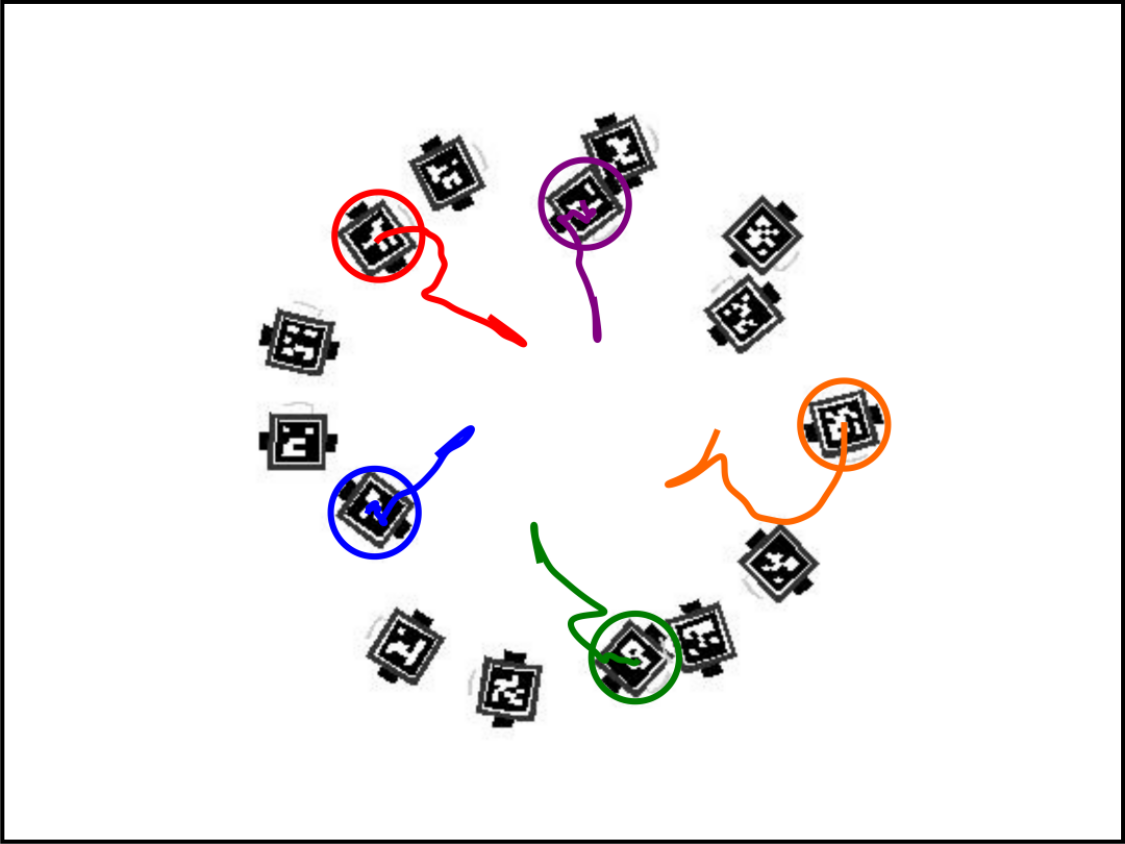}}%
	\caption{Sequence of snapshots of the \textit{surprise} behavior. The robots move along a circle of expanding radius, thus creating a spiral effect. The trajectories of five robots have been plotted using solid lines. See the full video at \url{https://youtu.be/VYIJ5hBeOIU}.}
	\label{fig:surpriseFrames}
\end{figure*}

\begin{figure*}[t!]
	\centering%
	\subfloat[\label{subfig:sadnessf1}$t =0$ s]{\includegraphics[width=0.33\linewidth]{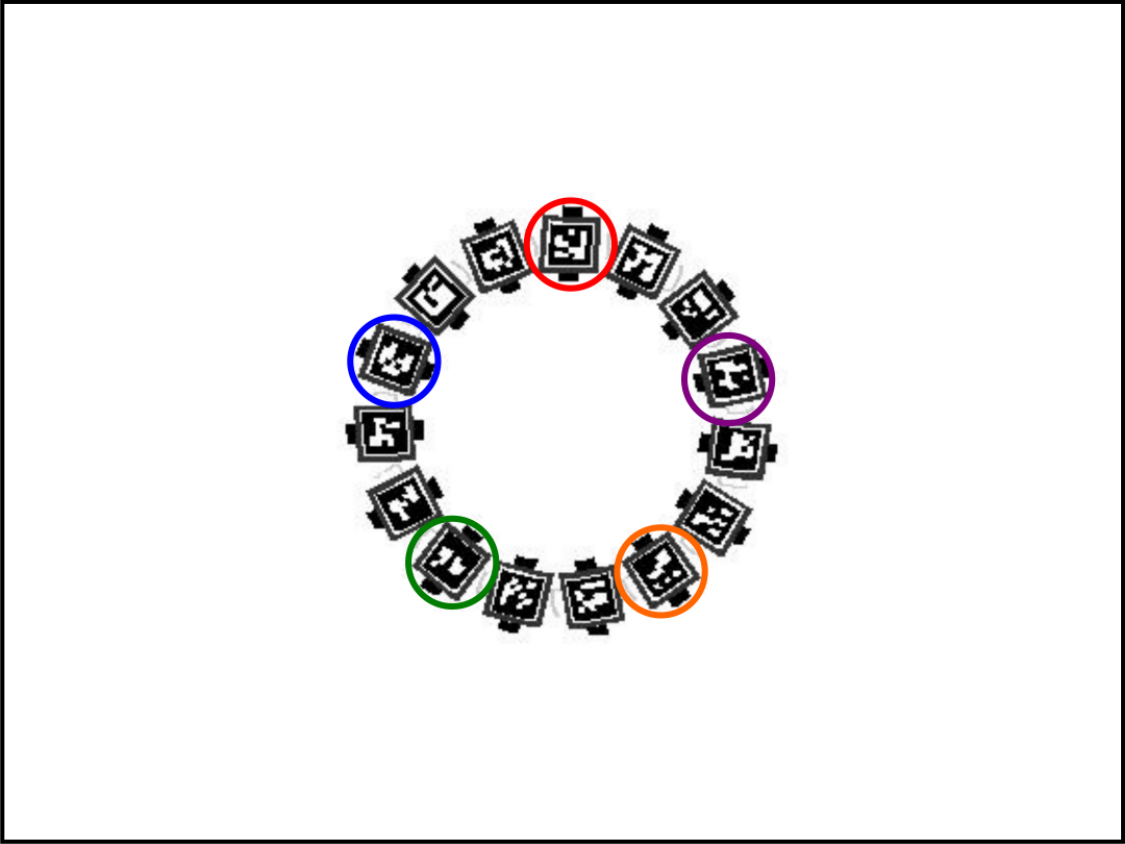}}%
	\subfloat[\label{subfig:sadnessf2}$t =2$ s]{\includegraphics[width=0.33\linewidth]{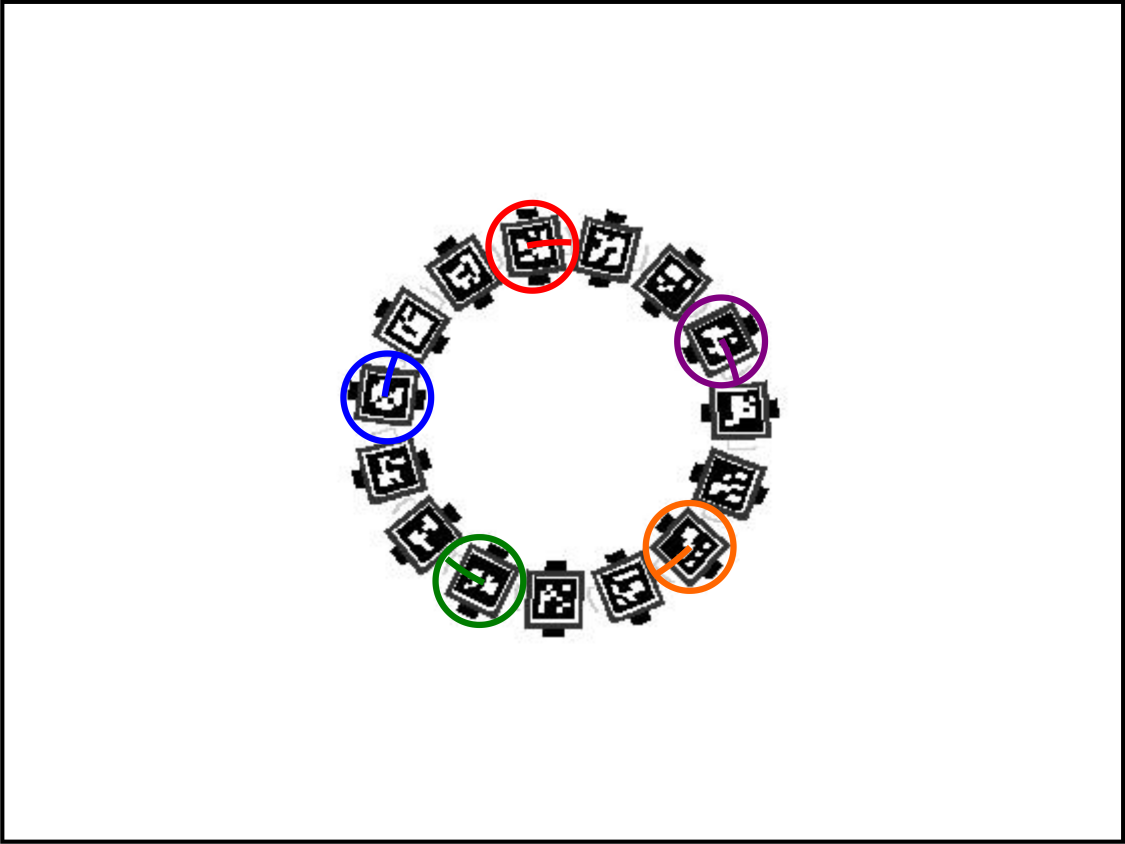}}%
	\subfloat[\label{subfig:sadnessf3}$t =8$ s]{\includegraphics[width=0.33\linewidth]{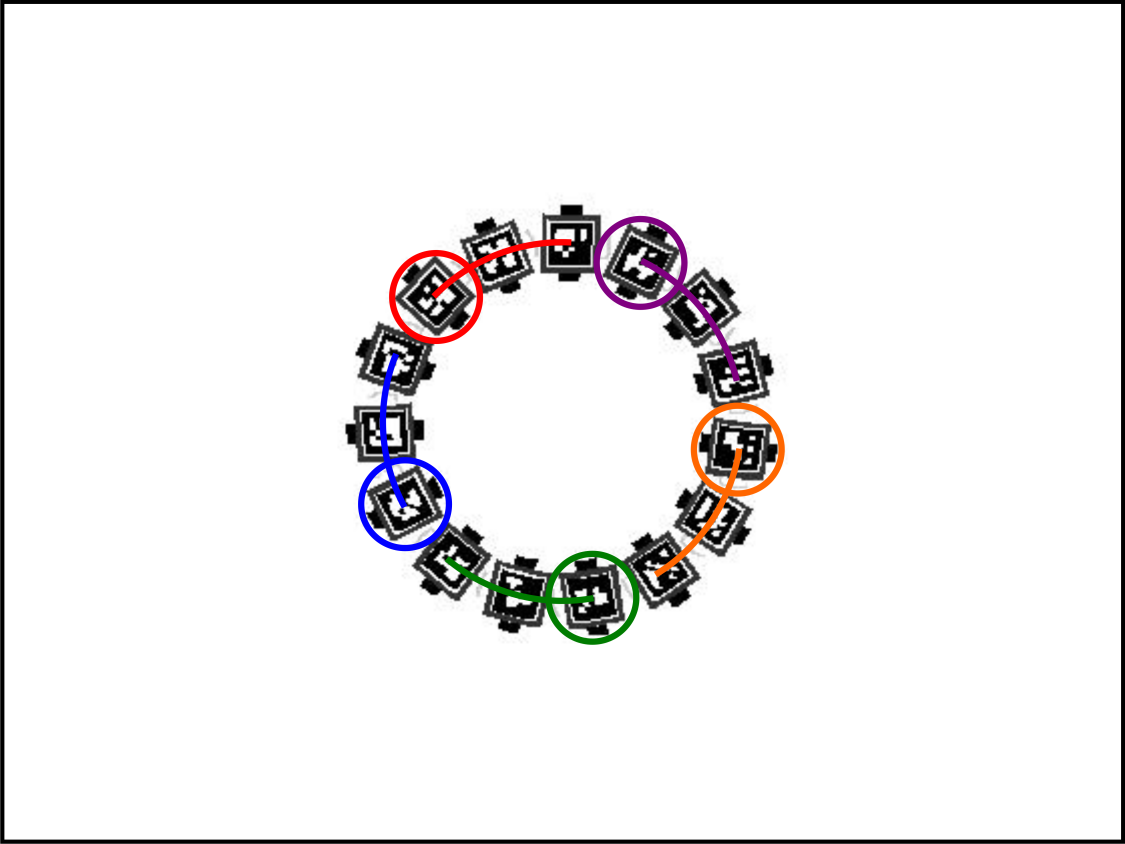}}%
	\caption{Sequence of snapshots of the \textit{sadness} behavior. The robots move along a small circle at a low speed. The trajectories of five robots have been plotted using solid lines. After 8 seconds, each robot has only displaced approximately an eighth of the circumference. See the full video at \url{https://youtu.be/rfHZcFnRFg8}.}
	\label{fig:sadnessFrames}
\end{figure*}

\begin{figure*}[t!]
	\centering%
	\subfloat[\label{subfig:fearf1}$t =0$ s]{\includegraphics[width=0.33\linewidth]{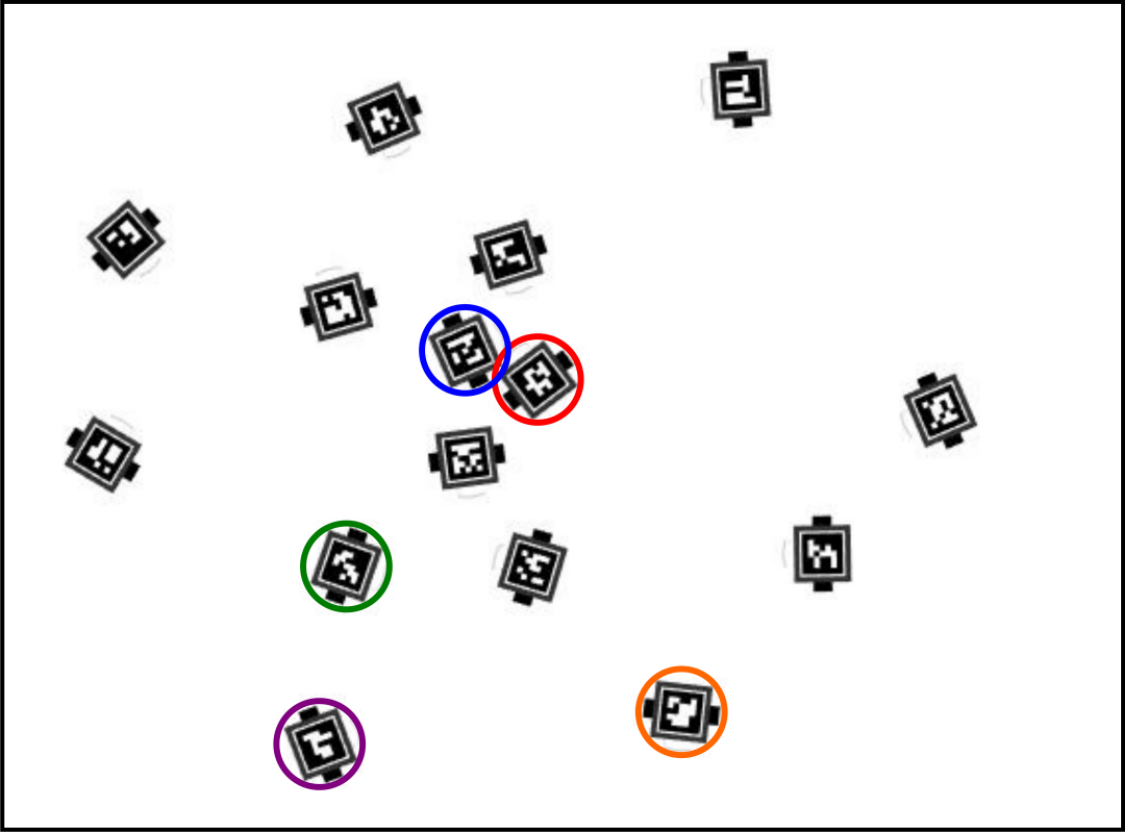}}%
	\subfloat[\label{subfig:fearf2}$t =3$ s]{\includegraphics[width=0.33\linewidth]{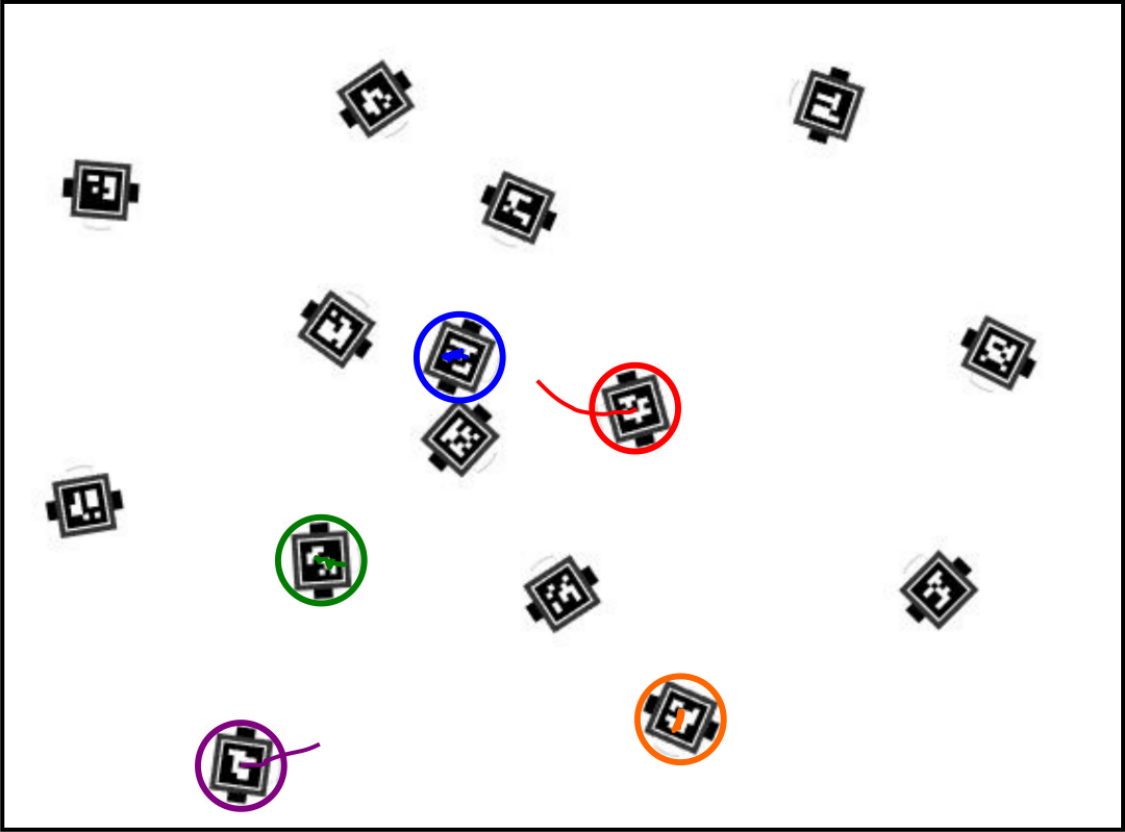}}%
	\subfloat[\label{subfig:fearf3}$t =15$ s]{\includegraphics[width=0.33\linewidth]{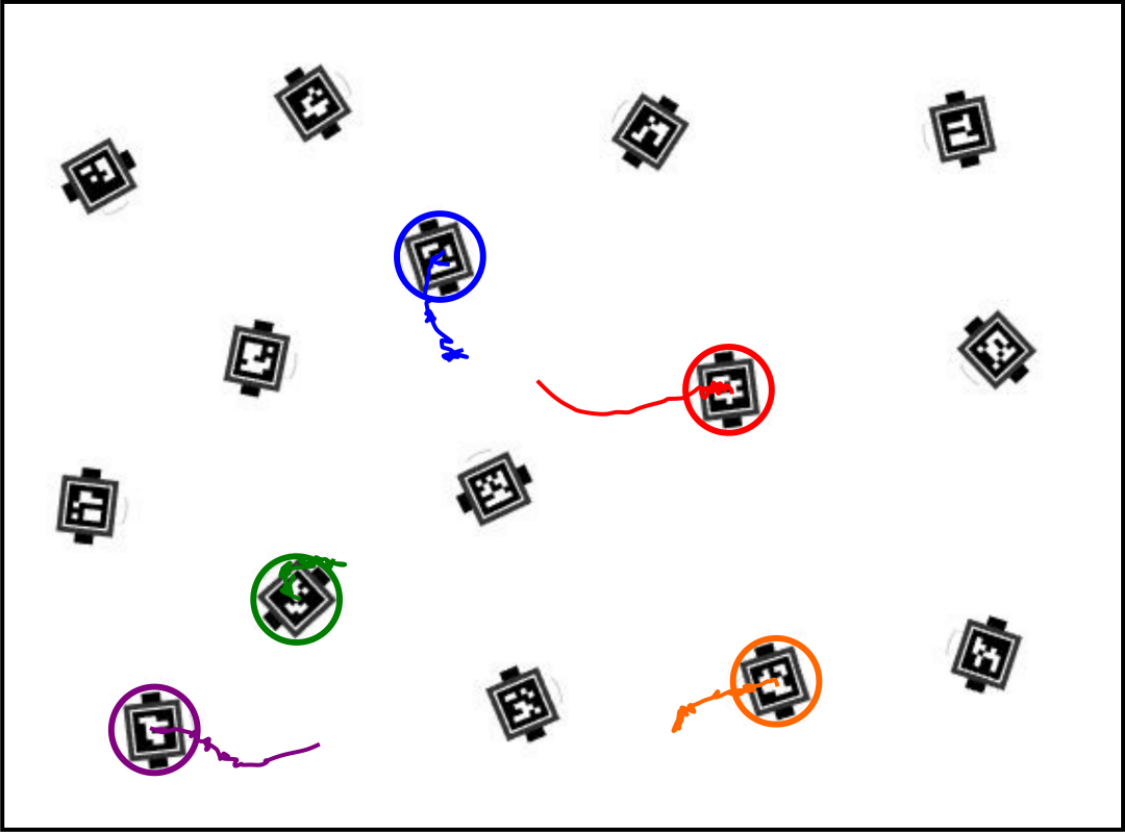}}%
	\caption{Sequence of snapshots of the \textit{fear} behavior. The robots spread out uniformly over the domain. As it can be observed from the trajectories, they displace slowly with a non-smooth, angular movement trace. See  the full video at \url{https://youtu.be/jz-5INUd8wc}.}
	\label{fig:fearFrames}
\end{figure*}

\begin{figure*}[t!]
	\centering%
	\subfloat[\label{subfig:disgustf1}$t =0$ s]{\includegraphics[width=0.33\linewidth]{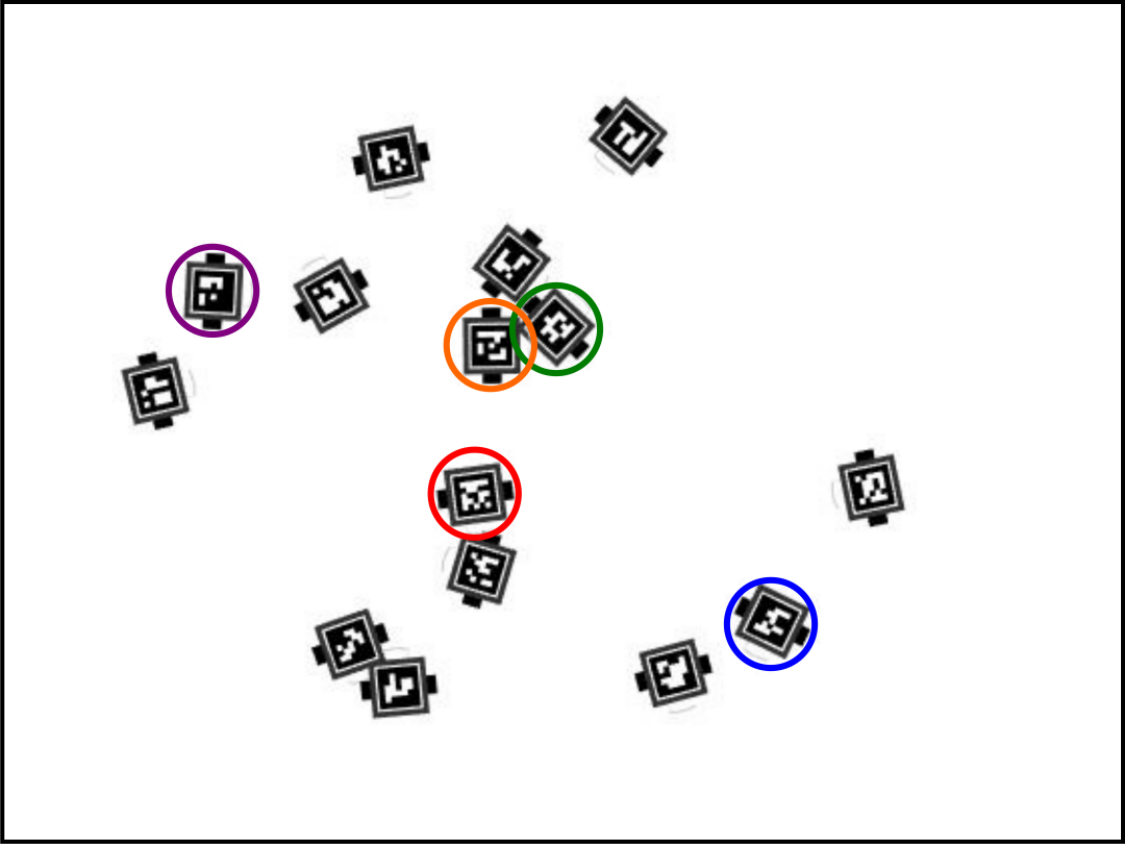}}%
	\subfloat[\label{subfig:disgustf2}$t =5$ s]{\includegraphics[width=0.33\linewidth]{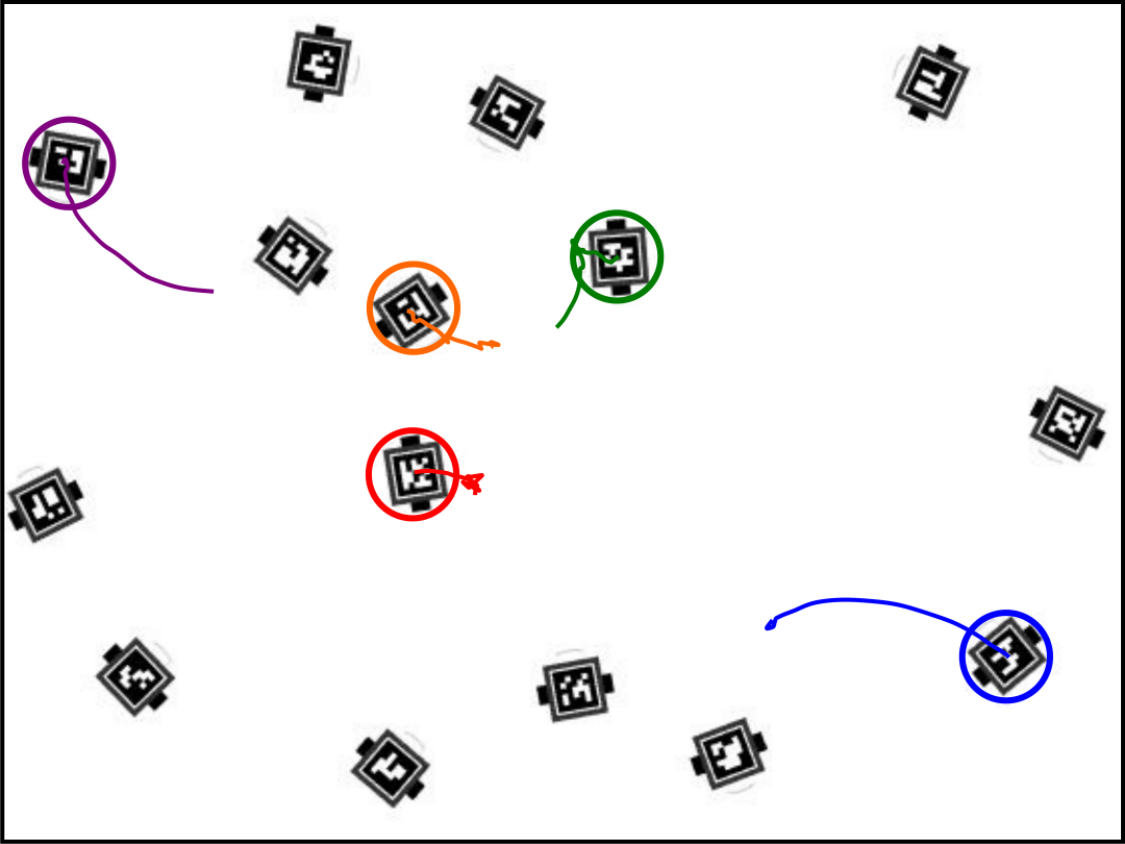}}%
	\subfloat[\label{subfig:disgustf3}$t =12$ s]{\includegraphics[width=0.33\linewidth]{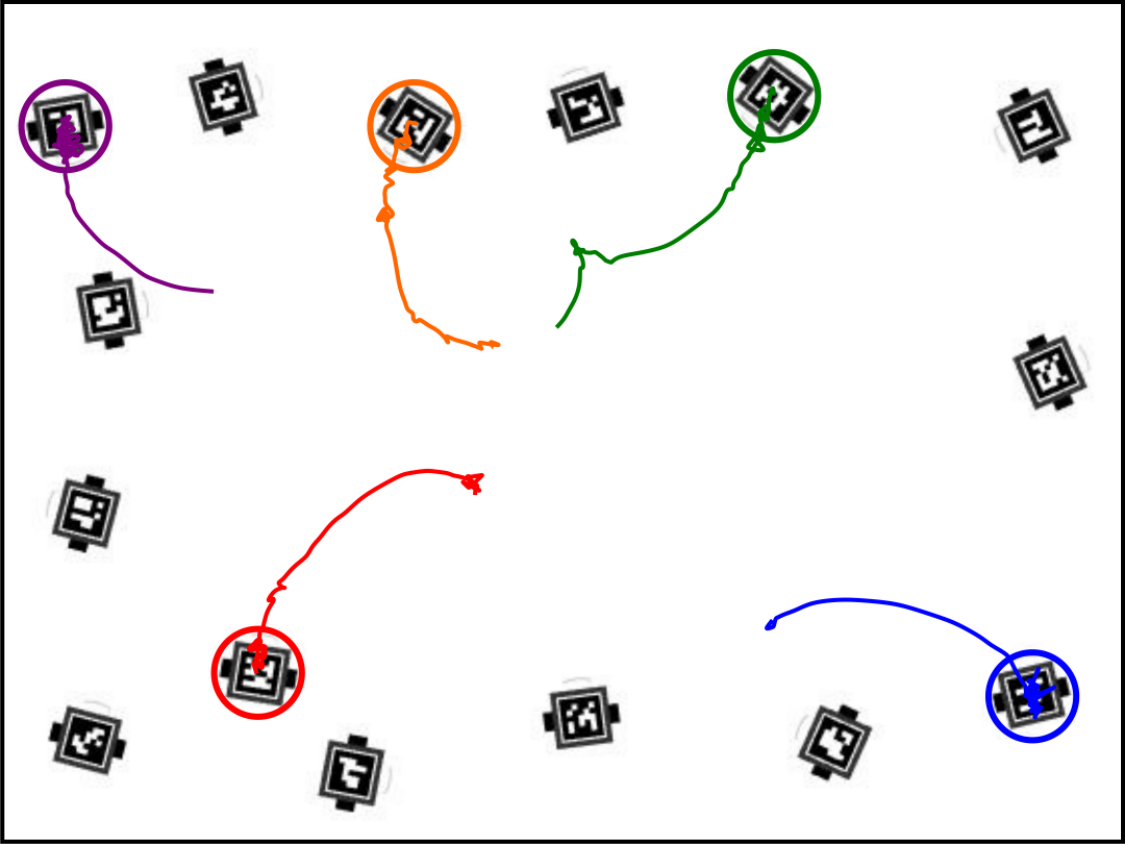}}%
	\caption{Sequence of snapshots of the \textit{disgust} behavior. The robots spread out slowly towards the boundaries of the domain, with a trajectory with a non-smooth, angular trace. See the full video at \url{https://youtu.be/EprfuCsuuRM}.}
	\label{fig:disgustFrames}
\end{figure*}

\begin{figure*}[t!]
	\centering%
	\subfloat[\label{subfig:angerf1}$t =0$ s]{\includegraphics[width=0.33\linewidth]{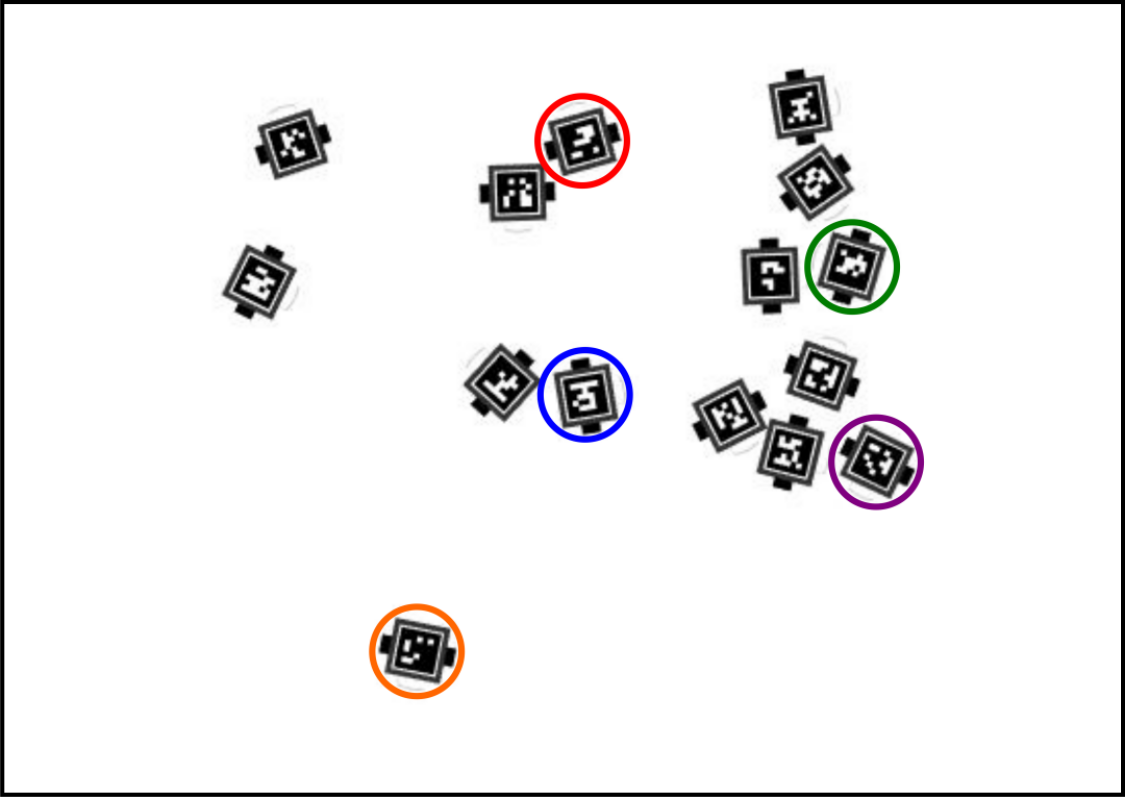}}%
	\subfloat[\label{subfig:angerf2}$t =2$ s]{\includegraphics[width=0.33\linewidth]{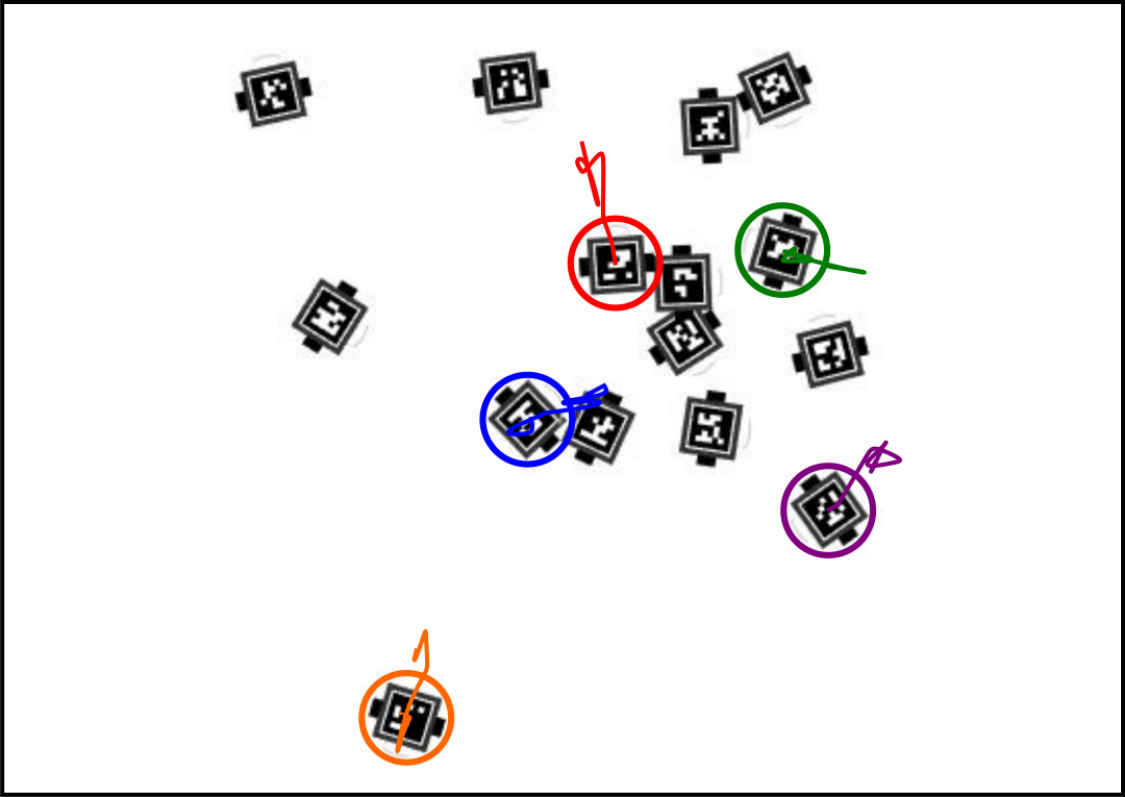}}%
	\subfloat[\label{subfig:angerf3}$t =6$ s]{\includegraphics[width=0.33\linewidth]{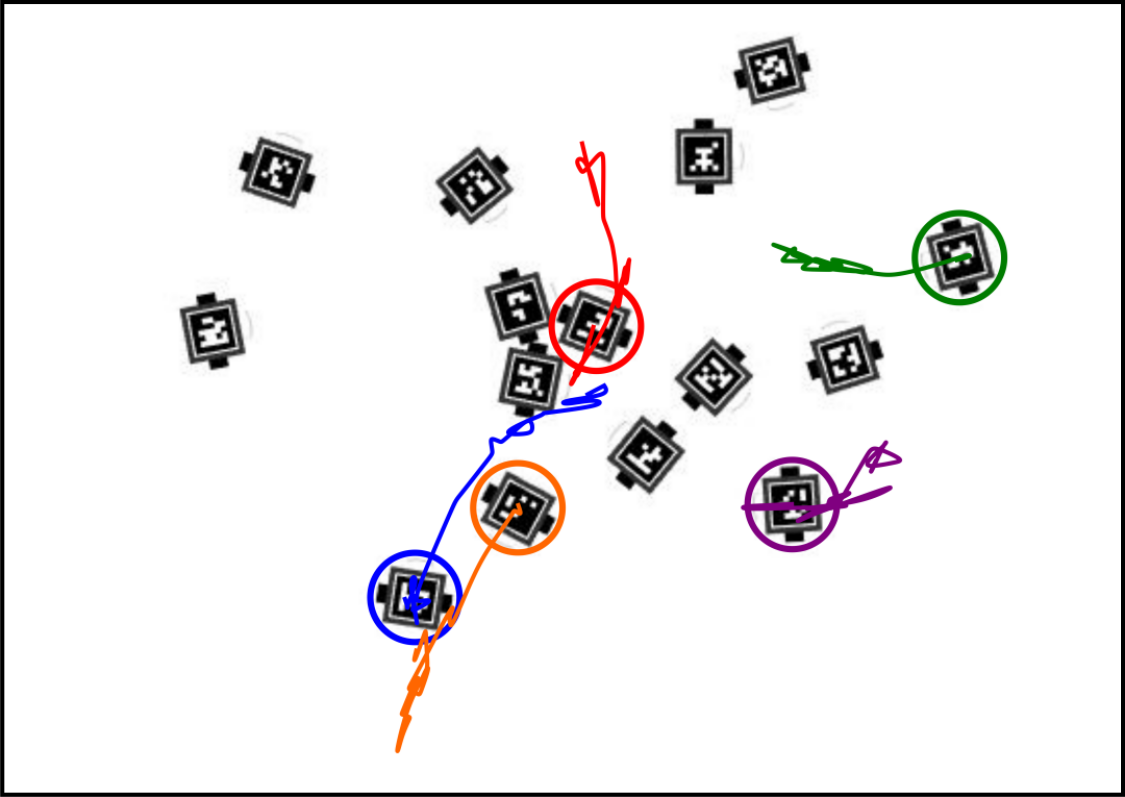}}%
	\caption{Sequence of snapshots of the \textit{anger} behavior. The density function is defined as a Gaussian at the center of the domain, causing the robots to concentrate around this area. However, the fact that the robots move with high speed causes overshoots in their positions, thus producing a significantly angular movement trace. See the full video at \url{https://youtu.be/kAGBrMkOtyY}.}
	\label{fig:angerFrames}
\end{figure*}

\subsection{Collective Behavior}
\label{sec:collective_behavior}
The attributes presented in Section \ref{sec:psychology} characterize how the motion and shape of an abstract object can convey emotion. Here we treat the GRITSBots as objects capable of reconfiguring themselves on a stage in order to generate an expressive behavior.

Among the attributes presented in Table \ref{tab:motion_shape_emotion}, it seems natural for those related to \textit{shape} and \textit{size} to be depicted by the collective behavior of the swarm, given that the individual robots can move within the planar environment but cannot change their individual shape. 
To this end, the feature of \textit{roundness} is incorporated into the behaviors of happiness, surprise and sadness. Those behaviors are thus based on the robots following some kind of circular contour, as illustrated in Figs. \ref{fig:happinessFrames}, \ref{fig:surpriseFrames} and \ref{fig:sadnessFrames}, respectively. In the case of the happiness behavior, a sinusoid is superimposed to the base shape of a circle, producing ripples on the circle contour to embody the \textit{curvilinearity} feature; and the corresponding size attribute---\textit{big}---is incorporated through the circle dimensions with respect to the domain. As for the surprise emotion, the \textit{very big} size attribute was included in the behavior by making the radius of the circle grow with time, thus producing a sensation of increasing size. Finally, the circular path dimension was reduced (\textit{small} attribute) in the case of the sadness behavior, incorporating also the \textit{slowness} attribute by making the robots follow the contour at a very low speed.

The scarcity of shape characterizations for the other three emotions---fear, disgust and anger---motivates a different approach for the design of the collective behavior of the swarm. For these emotions, we choose to specify which areas of the domain the robots should concentrate around. We do so by defining a density function, $\phi$, that characterizes the areas of the domain where we want the robots to group. In all three behaviors, the robots are initially distributed at random positions within the domain to then spread according to the particular density function selected. In the case of fear, the density function is uniform across the domain, so that it makes the robots scatter as far as possible from their neighbors, as shown in Fig. \ref{fig:fearFrames}. For the disgust motion, Fig. \ref{fig:disgustFrames}, the density is chosen to be high around the boundaries, making the robots move from the center towards the exterior of the domain---the stage---, giving the sensation of animosity between robots. Finally, in order to show anger, the robots are made to stay closer to the center of the domain. This strategy, combined with the individual robot control that will be explained in Section \ref{sec:individual_robot_control}, is intended to give the sensation of a heated environment, a riot.

The control laws needed to achieve these behaviors are explained in detail in Appendix \ref{sec:appendix_swarm_behaviors}. In each of those laws, a robot in the swarm is treated as a point that can move omnidirectionally. However, the GRITSBots (see Fig. \ref{fig:gritsbot}) are differential drive robots and, thus, are unable to move perpendicularly to the direction of their wheels. This movement restriction is used to our advantage in the individual control strategies described in Section \ref{sec:individual_robot_control}, where we exploit the limitations on the planar movement of the differential drive robots to implement the \textit{movement features} in Table  \ref{tab:motion_shape_emotion}.

\subsection{Individual Robot Control}
\label{sec:individual_robot_control}
The swarm behavior strategies and corresponding control laws introduced in Section \ref{sec:collective_behavior} and detailed in Appendix \ref{sec:appendix_swarm_behaviors} treat each robot in the swarm as if it could move omnidirectionally. That is, if we denote by $p\in\mathbb{R}^2$ the position of a robot, then its movement could be expressed using \textit{single integrator dynamics},
\begin{equation}\label{eq:single_integrator_dynamics}
 \dot p = u,
\end{equation}
with $u\in\mathbb{R}^2$ denoting the control action given by the chosen behavior. However, the differential drive configuration of the GRITSBot implies that it cannot execute single integrator dynamics. Instead, the motion of a differential drive robot is described by the so-called \textit{unicycle dynamics}, 
\begin{align}\label{eq:unicycle_dynamics_general}
 &\dot x = v \cos\theta,\nonumber\\
 &\dot y = v \sin\theta,\\
 &\dot \theta = \omega,\nonumber
\end{align}
with $p=(x,y)^T$ being the robot's cartesian position and $\theta$ its orientation in the plane. The control inputs, $v$ and $\omega$, correspond to the linear and angular velocities of the robot, respectively, as shown in Fig. \ref{fig:swarmVsIndividual}.

\begin{figure}[t!]
 \centering
 \includegraphics[width=\linewidth]{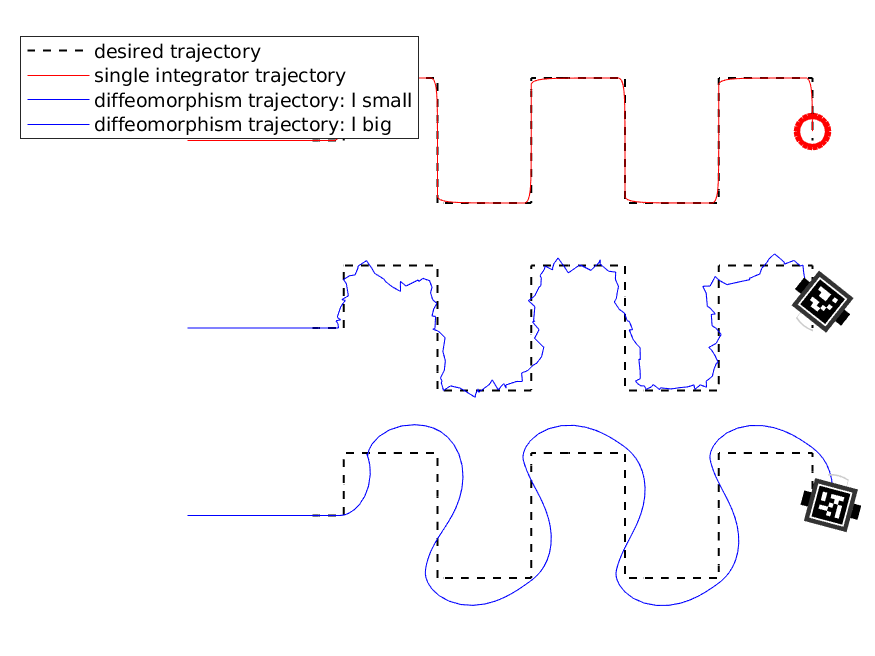}
 \caption{Effect of the diffeomorphism parameter, $l$, on the movement trace of an individual robot. In all cases, the controller is following a particle that moves along the black dashed line---the desired trajectory. The top figure illustrates how an agent capable of executing the single integrator dynamics in \ref{eq:single_integrator_dynamics} follows closely  the desired trajectory. The other two trajectories, in blue, illustrate two different diffeomorphisms performed over the control action of the single integrator. In the middle, a small value of $l$ results in an angular movement trace that follows quite closely the desired trajectory. In contrast, at the bottom, a large value of $l$ results on a very smooth movement trace, at the expense of following more loosely the desired trajectory.}
 \label{fig:diffeomorphism_trajectory}
\end{figure}

In order to convert the input $u$ in \eqref{eq:single_integrator_dynamics} into the executable control commands in \eqref{eq:unicycle_dynamics_general}, we use the near-identity diffeomorphism in \cite{Olfati-Saber2002}. The details of this transformation are described in detail in Appendix \ref{sec:appendix_diffeomorphism}. Using this transformation between the single integrator and the unicycle dynamics, we get to tune two scalar parameters, $l$ and $K$, that regulate how \textit{smooth} the movement trace of each robot is and how \textit{fast} it travels when executing a certain control input, respectively. Figure \ref{fig:diffeomorphism_trajectory} illustrates the differences between directly executing the single integrator dynamics in \eqref{eq:single_integrator_dynamics}, and performing two different diffeomorphisms on the single integrator control value, $u$. We can observe how choosing a small value for the diffeomorphism parameter $l$ results in an angular movement trace, while a smooth trajectory is observed when selecting a bigger value for this parameter.

Given the ability to regulate the angularity and the speed of the movement trace of a robot, we are in a position to implement the movement features included in  Table \ref{tab:motion_shape_emotion}. 
The \textit{smoothness} feature of the happiness emotion is translated into a smooth and fast individual control. Analogous diffeomorphism parameters are chosen to show surprise, given the \textit{roundness} and \textit{very big size} attributes associated with this emotion. As for sadness, even though it is a negative emotion, we focus on its specific characterizations provided in Table \ref{tab:motion_shape_emotion} to characterize the motion as slow and smooth. We can observe how, indeed, the trajectories depicted in Figs. \ref{fig:happinessFrames}, \ref{fig:surpriseFrames} and \ref{fig:sadnessFrames} are smooth given the choice of a large $l$ in the diffeomorphism. The speed of the robots is illustrated by the total distance covered in time: while significant distances are traveled within 4 seconds for the behaviors of happiness and surprise, the robots in the sadness behavior displace very little in 8 seconds.

Table \ref{tab:motion_shape_emotion} associates an  \textit{angular movement trace} with the emotions with negative valence. Consequently, a controller that produces an angular movement trace, corresponding to a small $l$ in the diffeomorphism, is selected for the remaining emotions---fear, disgust and anger. The  \textit{movement features} presented in Table \ref{tab:motion_shape_emotion} for anger and fear are translated into fast and slow control, respectively. Given the lack of characterization for the speed of disgust, we opt to implement a slow motion. We can observe how, for Figs. \ref{fig:fearFrames}-\ref{fig:angerFrames}, the trajectory traces have sharp turns and angularities, specially in the case of the anger behavior, which is accentuated by the proportional gain corresponding to a large velocity.

The swarm behavior selected for each of the emotions according to the shape characterizations discussed in Section \ref{sec:collective_behavior} and the diffeomorphism parameters in this section are summarized in Table \ref{tab:attributes_emotions}. 

\begin{table}
\caption{Motion and shape attributes selected for the behaviors associated with the fundamental emotions.}\label{tab:attributes_emotions}       
\renewcommand*{\arraystretch}{1.2}
\centering
\begin{tabular}{lcc}
\hline\noalign{\smallskip}
Emotion & Swarm Behavior & Robot Control \\
\noalign{\smallskip}\hline\noalign{\smallskip}
Happiness & sinusoid over circle & fast, smooth \\
Surprise & expanding circle & fast, smooth \\
Sadness & small circle & very slow, smooth\\
Fear & uniform coverage
& slow, angular\\
Disgust & coverage on boundaries & slow, angular\\
Anger & coverage on center
& fast, angular\\
\noalign{\smallskip}\hline
\end{tabular}
\end{table}

\begin{table*}[t!]
\centering
\caption{Confusion matrix calculated with the survey responses.}
\label{tab:confusionMatrix}       
\renewcommand*{\arraystretch}{1.1}
\begin{tabular}{clC{2cm}C{2cm}C{2cm}C{2cm}C{2cm}C{2cm}}
\hline\noalign{\smallskip}
& & \multicolumn{6}{c}{Signaled Emotion}\\
& & Happiness & Surprise & Anger & Fear & Disgust & Sadness \\
\noalign{\smallskip}\hline\noalign{\smallskip}
& Happiness & \bf{64.44} & 17.78 & 8.89 & 4.44 & 4.44 & 13.33\\
& Surprise & 11.11 & \bf{57.78} & 8.89 & 2.22 & 0.00 & 0.00\\
& Anger & 8.89 & 0.00 & \bf{55.56} & 13.33 & 15.56 & 4.44\\
& Fear & 6.67 & 13.33 & 20.00 & \bf{40.00} & 35.56 & 15.56\\
& Disgust & 6.67 & 4.44 & 4.44 & 26.67 & \bf{40.00} & 2.22\\
\rot{\rlap{~Response (\%)}}
& Sadness & 2.22 & 6.67 & 2.22 & 13.33 & 4.44 & \bf{64.44}\\
\noalign{\smallskip}\hline
\end{tabular}
\end{table*}

\section{User Study}
\label{sec:user_study}
The behaviors described in Section \ref{sec:swarm_behavior_design} were implemented in simulation on a team of 15 differential drive robots, producing a video for each of the emotions. Snapshots generated from each of the videos, along with the URL links, are included in Figs. \ref{fig:happinessFrames} to \ref{fig:angerFrames}. 

\subsection{Procedure}

A user study was conducted to evaluate if the swarm interactions and individual robot control strategies selected in Section \ref{sec:swarm_behavior_design} produce expressive swarm behaviors that correspond to the fundamental emotions. The hypothesis to test was the following,
\begin{description}
 \item [H1: Overall Classification.] Participants will perform better than chance in identifying the fundamental emotion each swarm behavior is intended to represent.
\end{description}

A total of 45 subjects (32 males and 13 females) participated in the study, with 29 of them not having any academic or professional background in robotics. As for the age of the participants, the distribution was as follows: 31.1\% between 18 and 24 years old, 60.0\% between 25 and 34 years old, 6.7\% between 35 and 44 years old, and 2.2\% between 45 and 54 years old. After responding to the demographic questions, each subject was shown 6 videos, each of them corresponding to the behaviors designed for each of the fundamental emotions. The videos were shown sequentially, one behavior at a time, and in a random order. The human subjects were instructed to watch each video in full, after which they were presented with a multiple choice (single answer) question to select the emotion that \textit{best described the movement of the robots in the video}, with the possible answers being the 6 fundamental emotions. The participants had no time limit when classifying the videos and were allowed to rewatch them as many times as desired. Furthermore, at any point, the participants were allowed to navigate to previous questions in the survey and modify their answers, if desired, before submitting the survey responses.

\subsection{Results and Discussion}
\label{sec:results}
\begin{figure}[t!]
 \centering
 \includegraphics[width=0.8\linewidth]{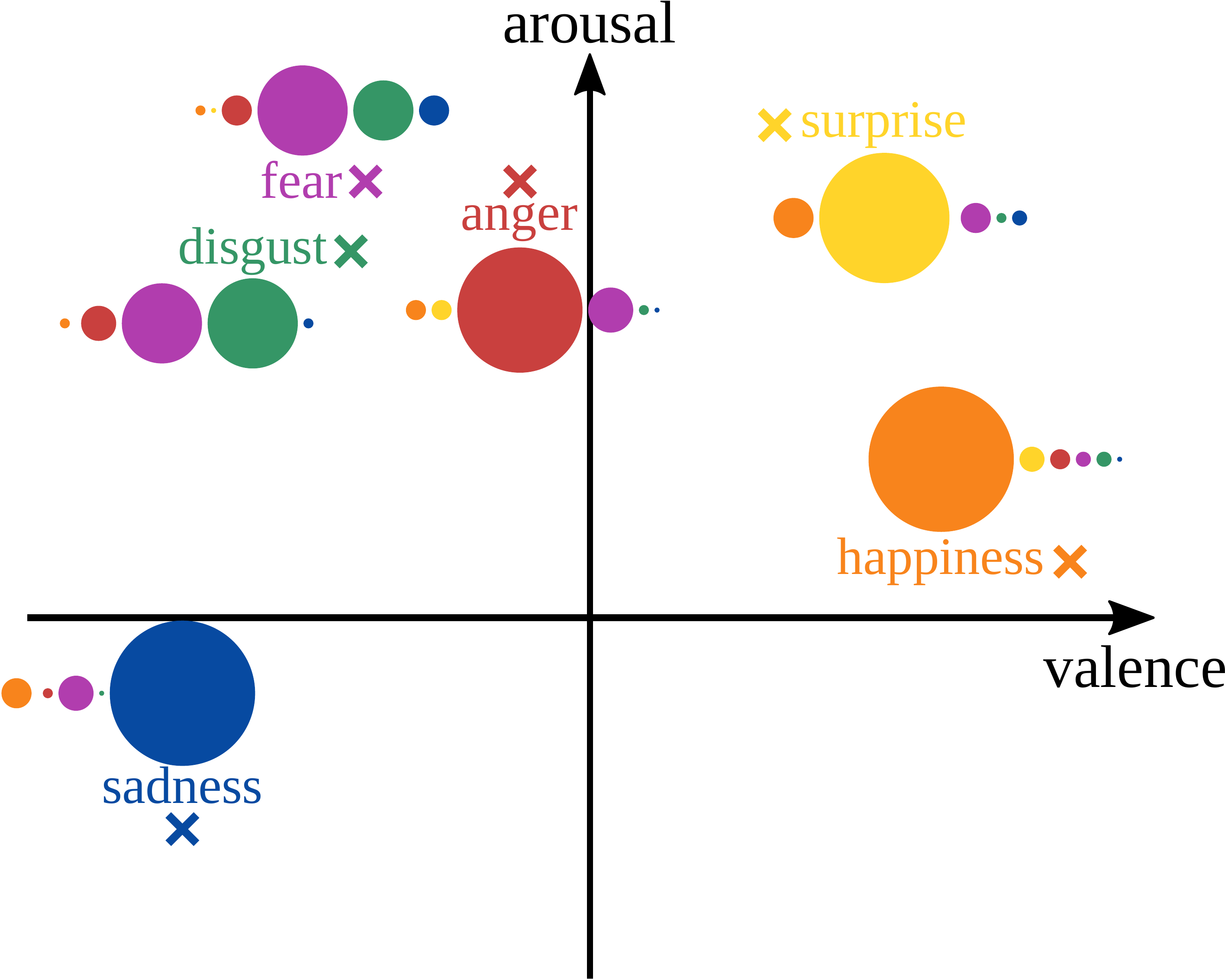}
 \caption{Representation of the survey responses in the valence-arousal plane. The location of each emotion is represented with a color-coded cross according to the circumplex model of affect \cite{Ross1938,Russell1980}. Next to each emotion, a sequence of color-coded circles represent how the human subjects identify each behavior, with the diameter of each circle being proportional to the amount of responses given to the corresponding emotion. We can observe how, in general, the majority of users labels the behavior according to the signaled emotion, with most variations occurring generally with those emotions closest in the plane. In the cases of fear and disgust, while the relative majority of subjects still labels their behaviors according to the hypothesis, we observe a significant amount of confusion among them, which may be due to the proximity of such emotions in terms of valence and arousal.}
 \label{fig:circumplex}
\end{figure}

The responses of the survey were collected and summarized in Table \ref{tab:confusionMatrix}. The columns are labeled \textit{signaled emotion} and each of them contains the responses given to the video of the behavior designed for a fundamental emotion. In the confusion matrix in Table \ref{tab:confusionMatrix}, the emotions are ordered counterclockwise from positive to negative valence according to the circumplex model in Fig. \ref{fig:circumplex}.

\begin{figure}[b!]
 \centering
 \includegraphics[width=\linewidth]{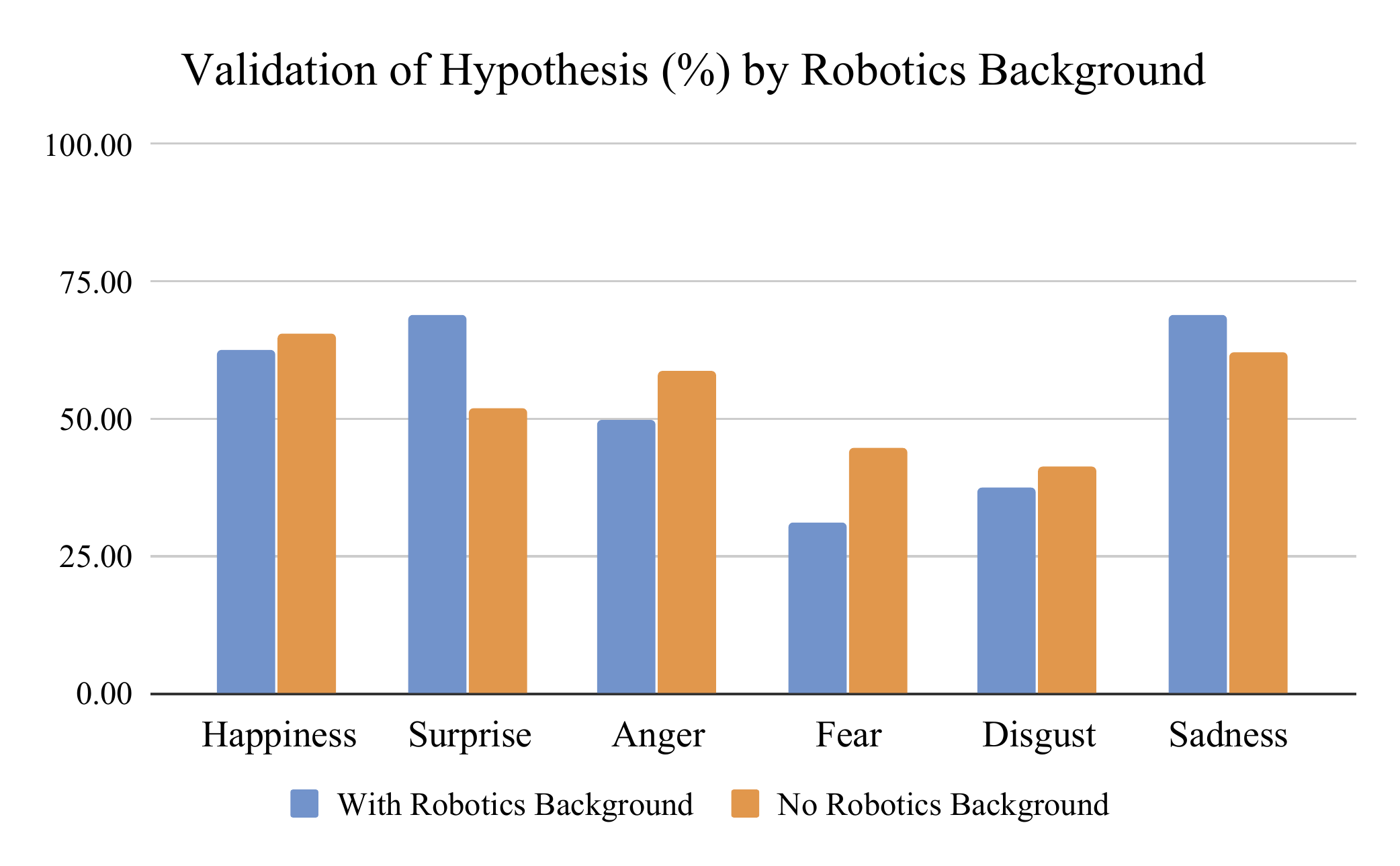}
 \caption{Percentage of subjects that identified each emotion in the video according to the hypothesis, classified according the robotics background of the subjects. There is no substantial difference between the responses given by the subjects that had experience studying or researching in robotics and those who did not.}
 \label{fig:chart_robotics}
\end{figure}

\begin{figure}[b!]
 \centering
 \includegraphics[width=\linewidth]{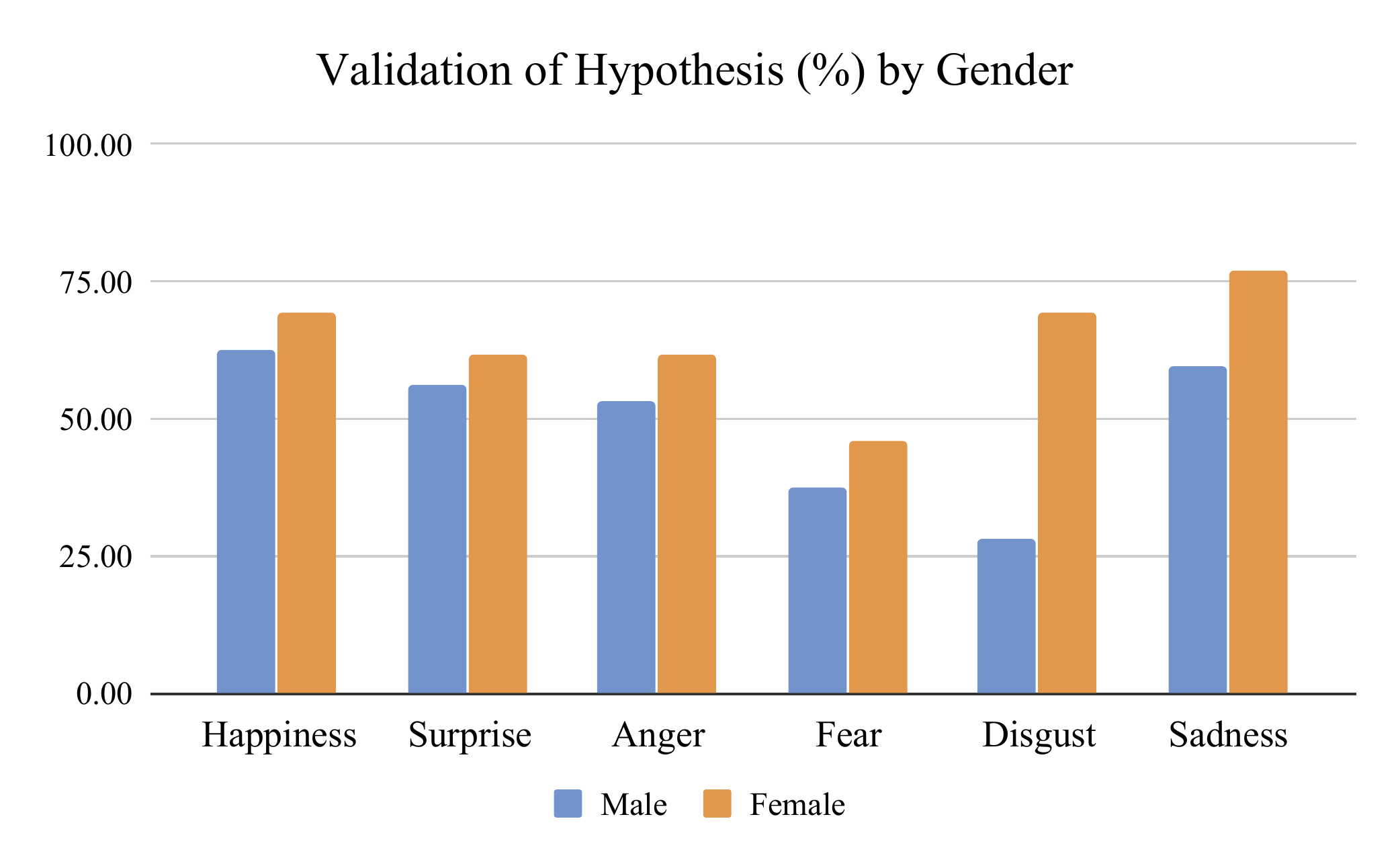}
 \caption{Percentage of subjects that successfully assigned the emotion to the corresponding video, according to the hypothesis, according to the gender of the participants. We can observe how the responses of the female subjects are consistently more aligned with the hypothesized behavior for each of the videos.}
 \label{fig:chart_gender}
\end{figure}

The diagonal terms of the confusion matrix, boldfaced in Table \ref{tab:confusionMatrix}, correspond to the percentage of responses that identified the emotion in the video as the one intended by the authors. For all the diagonal values, the percentage is much higher than the one given by chance (16.67\%), and in most cases---happiness, sadness, anger and surprise---this value reaches the absolute majority (greater than 50\%). In the cases of fear and disgust, while the relative majority of the responses identified the emotion according to our hypothesis (40\% for both emotions), the values are lower than 50\%. This can be potentially caused by the proximity of such emotions in terms of valence and arousal, as illustrated in Fig. \ref{fig:circumplex}. A Pearson's chi-squared test goodness of fit was performed for the responses given to each swarm behavior, confirming that, at $p<0.0001$, the frequency distributions for each emotion differ significantly with respect to a uniform distribution where all the emotions are considered equally likely to be chosen. Therefore, the assignment of an emotion to each of the videos  was not made at random by the participants, but rather the movement and shape features incorporated in the swarm behaviors were effectively identified as the intended emotions. 

\begin{figure*}[t!]
	\centering%
	\subfloat[][\label{subfig:happinessOverhead}
	\begin{tabular}{c}
	Happiness\\	
	\url{https://youtu.be/HQ6YkoADMBg}.
	\end{tabular}]
	{\includegraphics[width=0.33\linewidth]{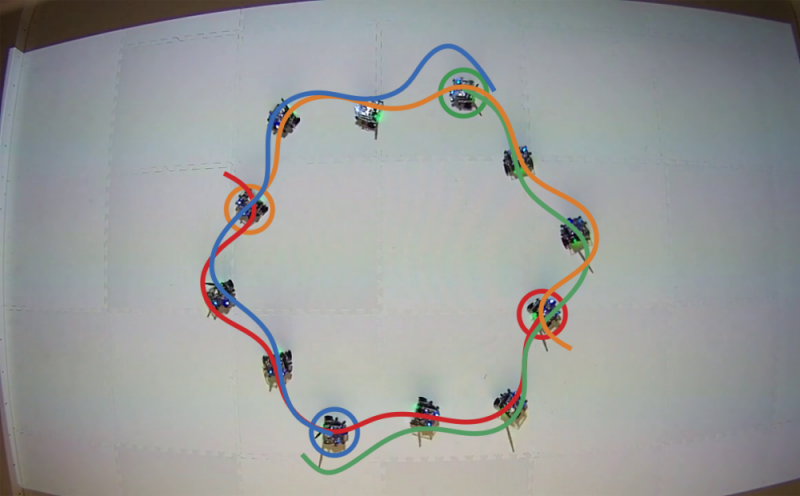}}
	\hfill
	\subfloat[][\label{subfig:surpriseOverhead}
	\begin{tabular}{c}
	Surprise\\
	\url{https://youtu.be/xhPTQg4iLvM}.
	\end{tabular}]{\includegraphics[width=0.33\linewidth]{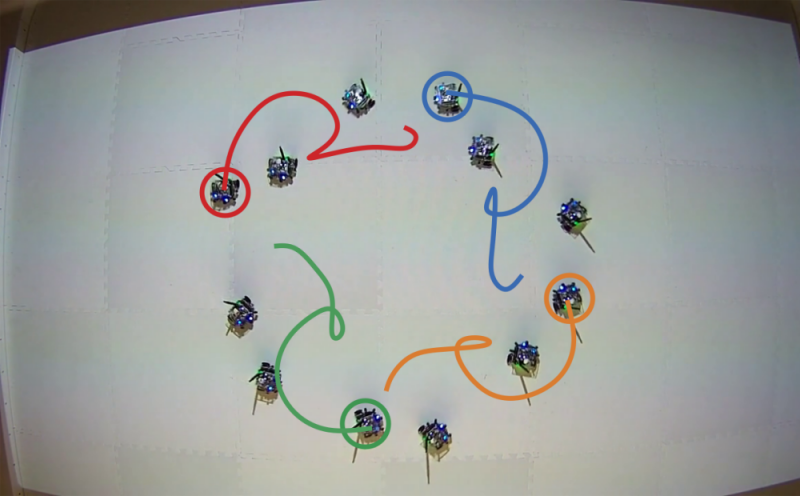}}
	\hfill
	\subfloat[][\label{subfig:sadnessOverhead}
	\begin{tabular}{c}
	Sadness\\	
	\url{https://youtu.be/i7cLP_GcL54}.
	\end{tabular}
	]{\includegraphics[width=0.33\linewidth]{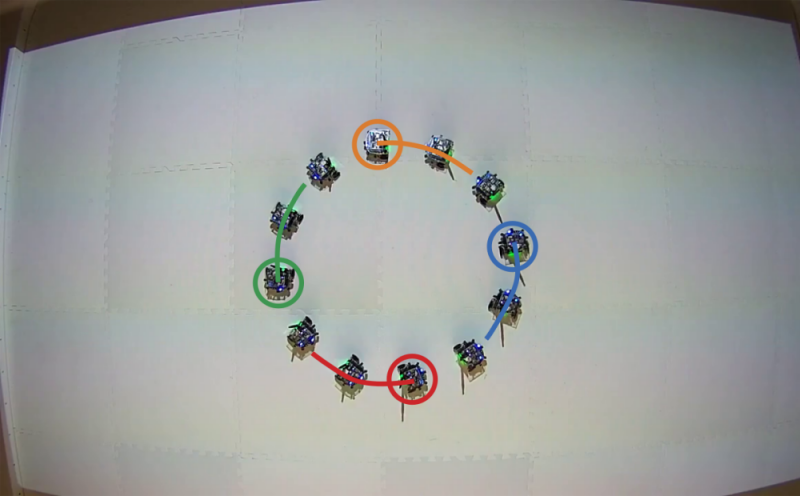}}\\
	\subfloat[][\label{subfig:fearOverhead}
	\begin{tabular}{c}
	Fear\\	
	\url{https://youtu.be/6xqb-sQck6I}.
	\end{tabular}
	]{\includegraphics[width=0.33\linewidth]{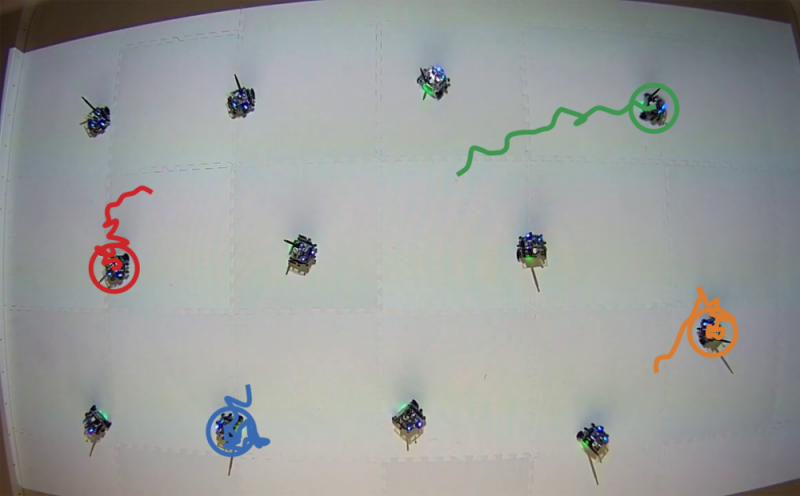}}%
	\hfill
	\subfloat[][\label{subfig:disgustOverhead}
	\begin{tabular}{c}
	Disgust\\	
	\url{https://youtu.be/RgPyXVuprX8}.
	\end{tabular}
	]{\includegraphics[width=0.33\linewidth]{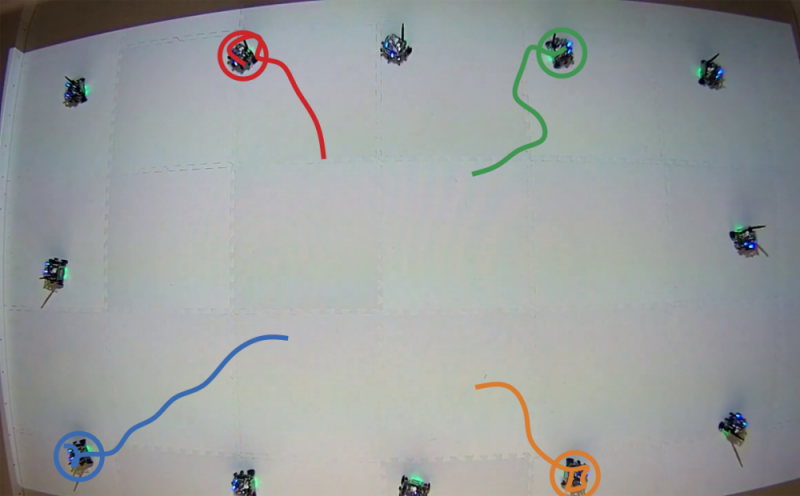}}	
	\hfill
	\subfloat[][\label{subfig:angerOverhead}
	\begin{tabular}{c}
	Anger\\	
	\url{https://youtu.be/VGlLPJGlwvo}.
	\end{tabular}
	]{\includegraphics[width=0.33\linewidth]{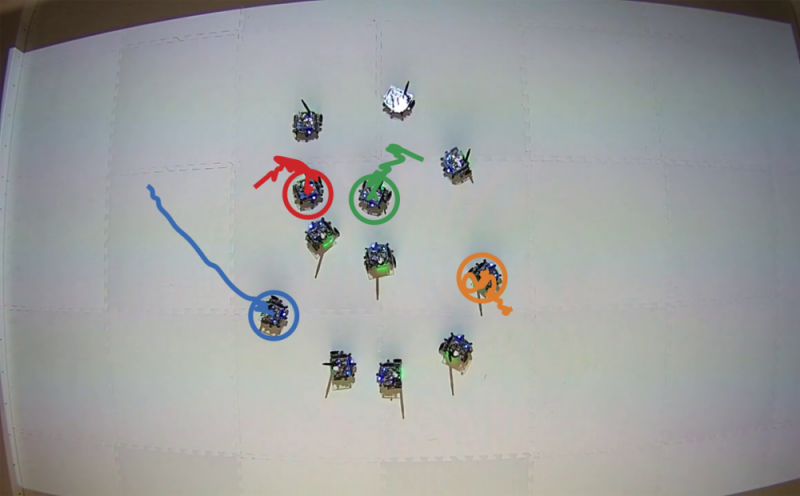}}
	\caption{Snapshots of the swarm behaviors implemented on a team of 12 GRITSBot X, taken in the Robotarium with an overhead camera that provides an analogous perspective to the one used in the simulations (Figs. \ref{fig:happinessFrames} to \ref{fig:angerFrames}). The trajectories of four robots have been plotted using solid lines. A link to the full video of each behavior is provided below each snapshot.}
	\label{fig:robotarium_overhead_frames}
\end{figure*}

Based on the demographic data collected, the validation of hypothesis H1 was not affected significantly by the robotics background of the subjects. As shown in Fig. \ref{fig:chart_robotics}, for the 4 emotions for which the majority of the aggregate responses in Table \ref{tab:confusionMatrix} aligned with the hypothesis---i.e. happiness, surprise, anger and sadness---all subjects, regardless of their background in robotics, identified the emotions according to the hypothesis in more than 50\% of the cases. In fact, the Pearson's chi-square test discards, at $p<0.01$, the random assignment of emotions from the responses of participants both with and without robotics background. For the emotions of fear and disgust---those with the lowest accuracies in Table \ref{tab:confusionMatrix}---the responses aligned better with hypothesis H1 for those subjects without a robotics background, for which the Pearson's chi-square test discards the fitting of the data under a uniform distribution at a significance level of $p<0.01$. While the subjects with robotics background still validated hypothesis H1 for these two emotions, the significance levels for the test are slightly higher ($p<0.05$ for fear and $p<0.1$ for disgust), possibly due to the fact that there were only 16 subjects with robotics background.

In contrast, when performing an analysis by gender, the validation of hypothesis H1 was consistently larger in the case of female subjects, as shown in Fig. \ref{fig:chart_gender}. While the male participants still validated hypothesis H1 for all emotions, the accuracy was higher among the female subjects, being in 5 out of the 6 emotions higher than 50\%. Only in the case of fear the accuracy for the female participants was slightly under the majority threshold (46.15\%). As for the statistical significance of the responses, the frequency of distributions for each emotion differs from a uniform distribution at $p<0.05$ for the male participants and at $p<0.01$ for the female ones. Thus, while neither of the populations assign emotions to the behaviors at random, the motion and shape characterizations selected for the swarm behaviors were more clearly identified by the female participants in the study.

In conclusion, the data collected in the user study unanimously supports hypothesis H1, thus confirming that the swarm behaviors and individual robot control paradigms designed in Section \ref{sec:swarm_behavior_design} effectively depict each of the fundamental emotions. Therefore, the behaviors considered in this study provide a collection of motion primitives for robotic swarms to effectually convey emotions in artistic expositions. 

\section{Robotic Implementation}
\label{sec:robotic_implementation}
\begin{figure*}[t!]
	\centering%
    \subfloat[][\label{subfig:happinessPerspective}
	\begin{tabular}{c}
	Happiness.\\	
	\url{https://youtu.be/EeEyIGn2BV0}.
	\end{tabular}]
	{\includegraphics[width=0.33\linewidth]{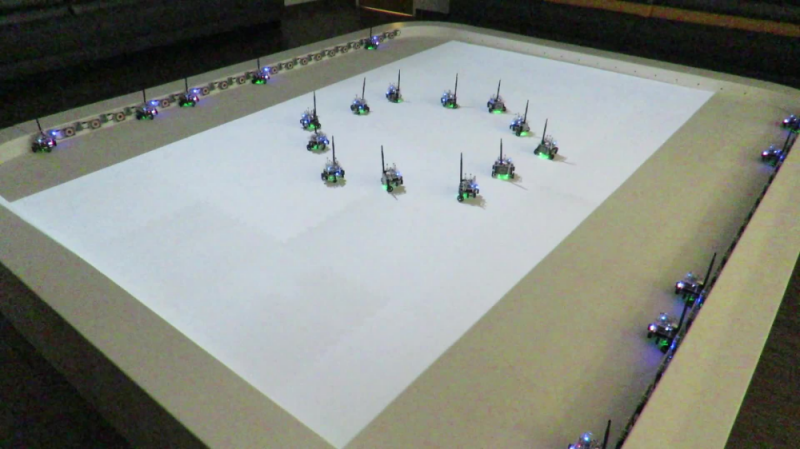}}
	\hfill
	\subfloat[][\label{subfig:surprisePerspective}
	\begin{tabular}{c}
	Surprise\\	
	\url{https://youtu.be/hHMjYMv6Ojo}.
	\end{tabular}]{\includegraphics[width=0.33\linewidth]{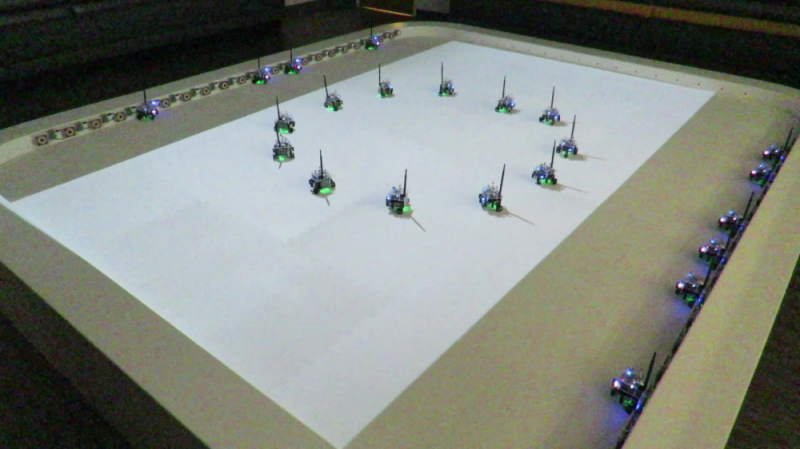}}
	\hfill
	\subfloat[][\label{subfig:sadnessPerspective}
	\begin{tabular}{c}
	Sadness\\	
	\url{https://youtu.be/jFWMtu5oYEo}.
	\end{tabular}
	]{\includegraphics[width=0.33\linewidth]{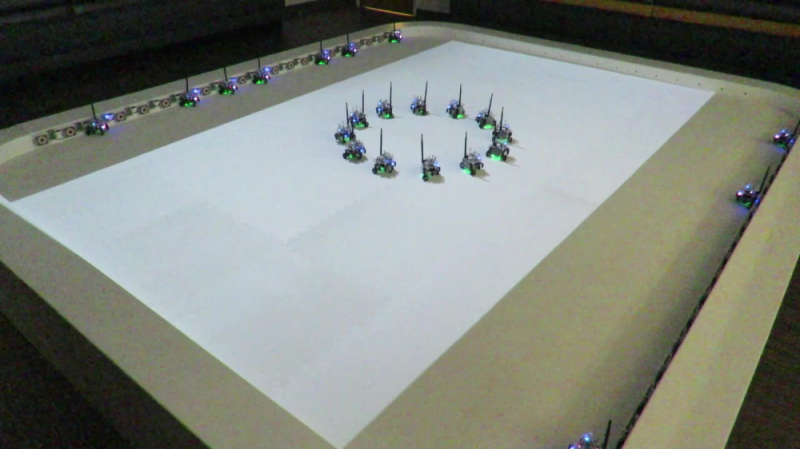}}\\
	\subfloat[][\label{subfig:fearPerspective}
	\begin{tabular}{c}
	Fear\\	
	\url{https://youtu.be/j72EXA14Scs}.
	\end{tabular}
	]{\includegraphics[width=0.33\linewidth]{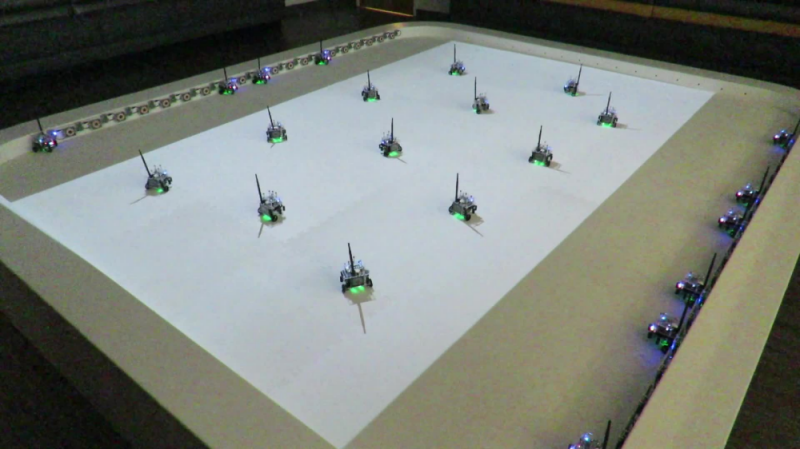}}%
	\hfill
	\subfloat[][\label{subfig:disgustPerspective}
	\begin{tabular}{c}
	Disgust\\	
	\url{https://youtu.be/py_cUXCkgZM}.
	\end{tabular}
	]{\includegraphics[width=0.33\linewidth]{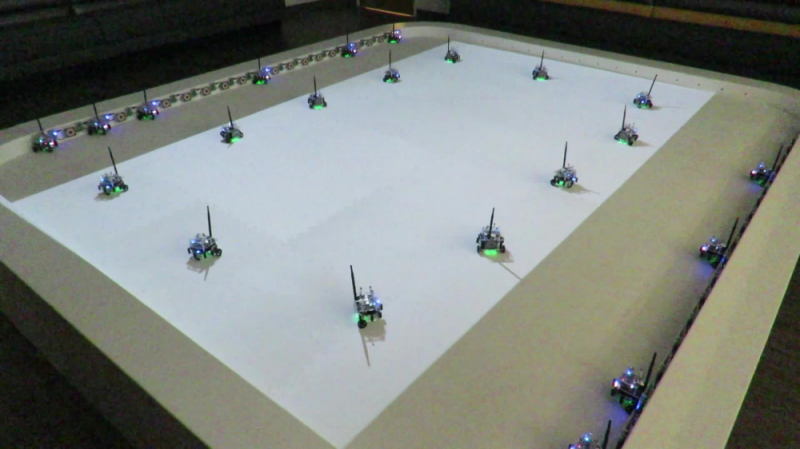}}	
	\hfill
	\subfloat[][\label{subfig:angerPerspective}
	\begin{tabular}{c}
	Anger\\	
	\url{https://youtu.be/Thj5s1vQvYA}.
	\end{tabular}
	]{\includegraphics[width=0.33\linewidth]{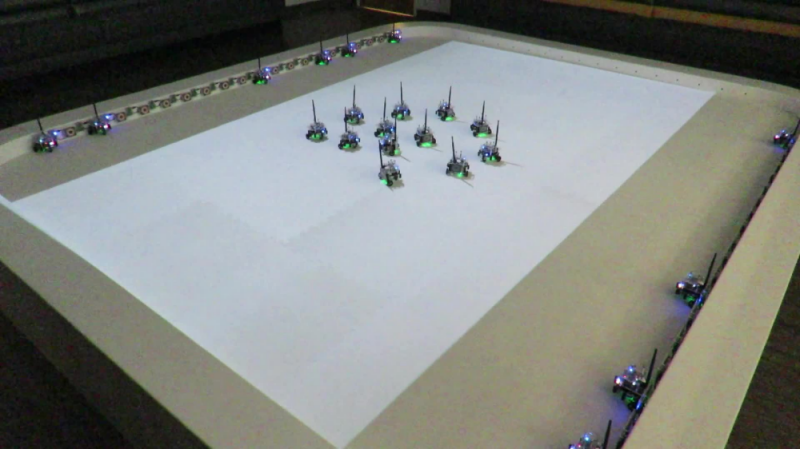}}
	\caption{Snapshots of the swarm behaviors implemented on a team of 12 GRITSBot X in the Robotarium, from a perspective point of view. The snapshots, taken with a camera located 1.70m over the Robotarium surface, provide a similar angle view to that of a human spectator. A link to the full video is provided for each behavior.}
	\label{fig:robotarium_perspective_frames}
\end{figure*}

The swarm behaviors proposed in Section \ref{sec:swarm_behavior_design} and simulated for the user study in Section \ref{sec:user_study} were implemented on a real robotic platform to evaluate their efficacy. Each behavior was executed by a team of 12 GRITSBots X on the Robotarium, a remotely accessible swarm robotics testbed at the Georgia Institute of Technology \cite{Wilson2020}. Similarly to the GRITSBot (images/Fig. \ref{fig:gritsbot}), the GRITSBot X has a differential-drive configuration, but with a bigger size: a 10cm$\times$10cm footprint. The robots move on the Robotarium arena, a 4.3m$\times$3.6m surface. The setup is shown in Figs. \ref{fig:robotarium_overhead_frames} and \ref{fig:robotarium_perspective_frames}.

The transition from the simulated behaviors in Section \ref{sec:swarm_behavior_design} and Appendices \ref{sec:appendix_swarm_behaviors} and \ref{sec:appendix_diffeomorphism} to their implementation on a real robotic platform involved the tuning of the parameters of the shapes and density functions associated with the behaviors, in accordance to the changes in size of the individual robots as well as of the Robotarium arena. Furthermore, the diffeomorphism parameters ($l$ and $K$ in Section \ref{sec:individual_robot_control}), while still reflected the specifications in Table \ref{tab:attributes_emotions} qualitatively, were adjusted to accommodate the dynamics and actuator limits of the GRITSBot X.

The resulting robotic behaviors are illustrated in Figs. \ref{fig:robotarium_overhead_frames} and \ref{fig:robotarium_perspective_frames}. Figure \ref{fig:robotarium_overhead_frames} presents a top view, analogous to the perspective used in the simulations (Figs. \ref{fig:happinessFrames} to \ref{fig:angerFrames}), with the purpose of showing the similarity between the simulated behaviors and the real behaviors. As can be observed in the snapshots and linked videos, for most emotions the simulated and real behavior do not present significant differences. The biggest contrast emerges for the anger emotion, where the actuator limits and safety constraints of the GRITSBot X prevent an exact replication of the simulated behavior, where very high peak velocities were executed by some individuals. Nevertheless, the behavior still portrays its characteristic features as described in Section \ref{sec:swarm_behavior_design}. A perspective view of the experiments taken at 1.70m over the Robotarium surface is presented in Fig. \ref{fig:robotarium_perspective_frames}. Despite changing the angle of view to that of an average person, the behaviors are still identifiable and highly distinctive.

\section{Conclusions}
\label{sec:conclusions}
In this paper, we investigated how motion and shape descriptors from social psychology can be integrated into the control laws of a swarm of robots to express fundamental emotions. Based on such descriptors, a series of swarm behaviors were developed, and their effectiveness in depicting each of the fundamental emotions was analyzed in a user study. The results of the survey showed that, for all the swarm behaviors created, the relative majority of the subjects classified each behavior with the corresponding emotion according to the hypothesis, being this ratio over 50\% for 4 of the 6 fundamental emotions.

\begin{figure*}[t!]
	\centering%
	\subfloat[\label{subfig:happinessShape}Happiness: The robots follow points moving along a circle of radius $R$ with a superposed sinusoid of amplitude $A$.]{\includegraphics[width=0.32\linewidth]{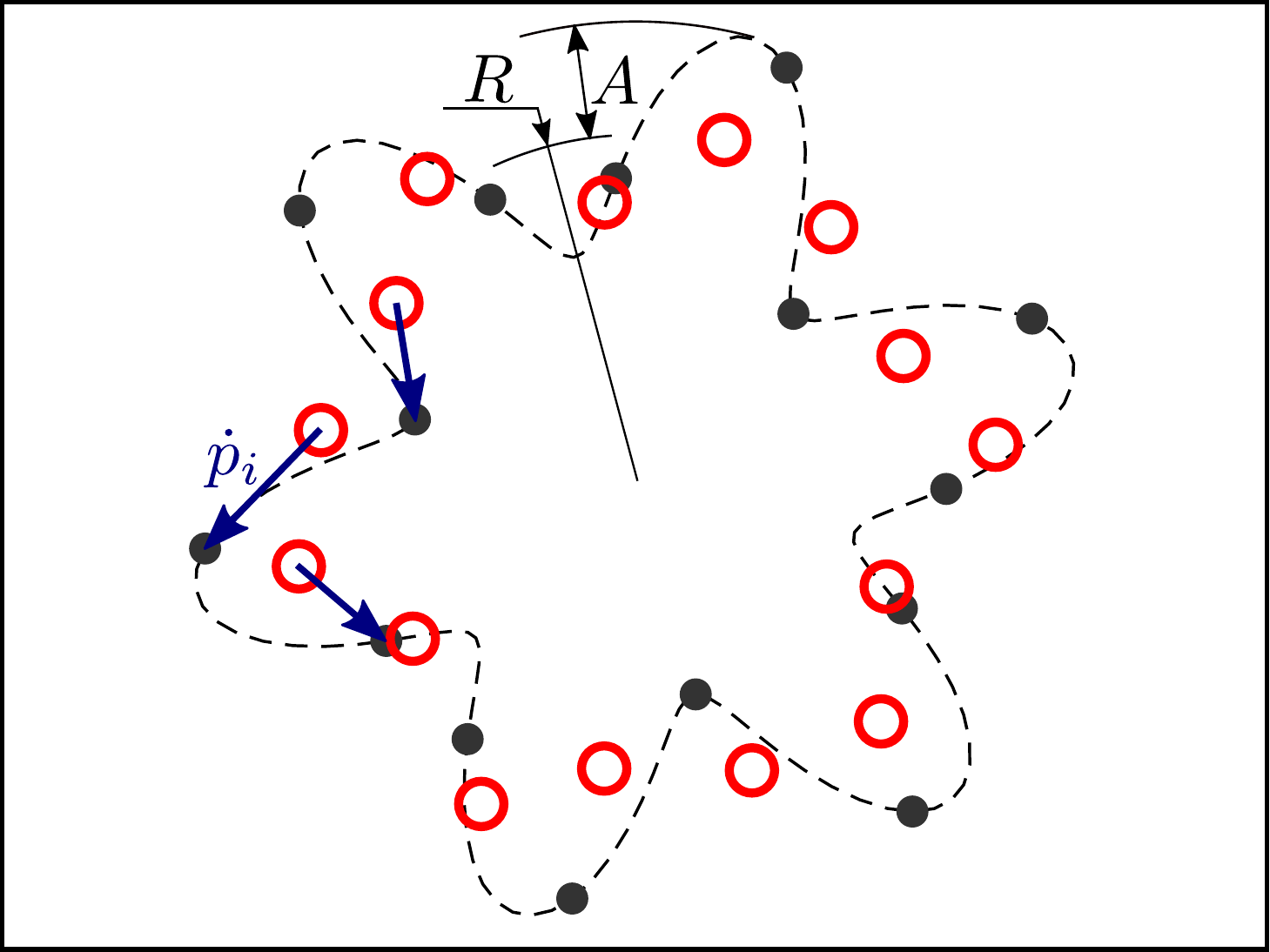}}%
	\hfill%
	\subfloat[\label{subfig:surpriseShape}Surprise: The robots follow points moving along a circle of expanding radius. Two snapshots, corresponding to $R(t)=\{R_{min}, R_{max}\}$, are shown here.]{\includegraphics[width=0.32\linewidth]{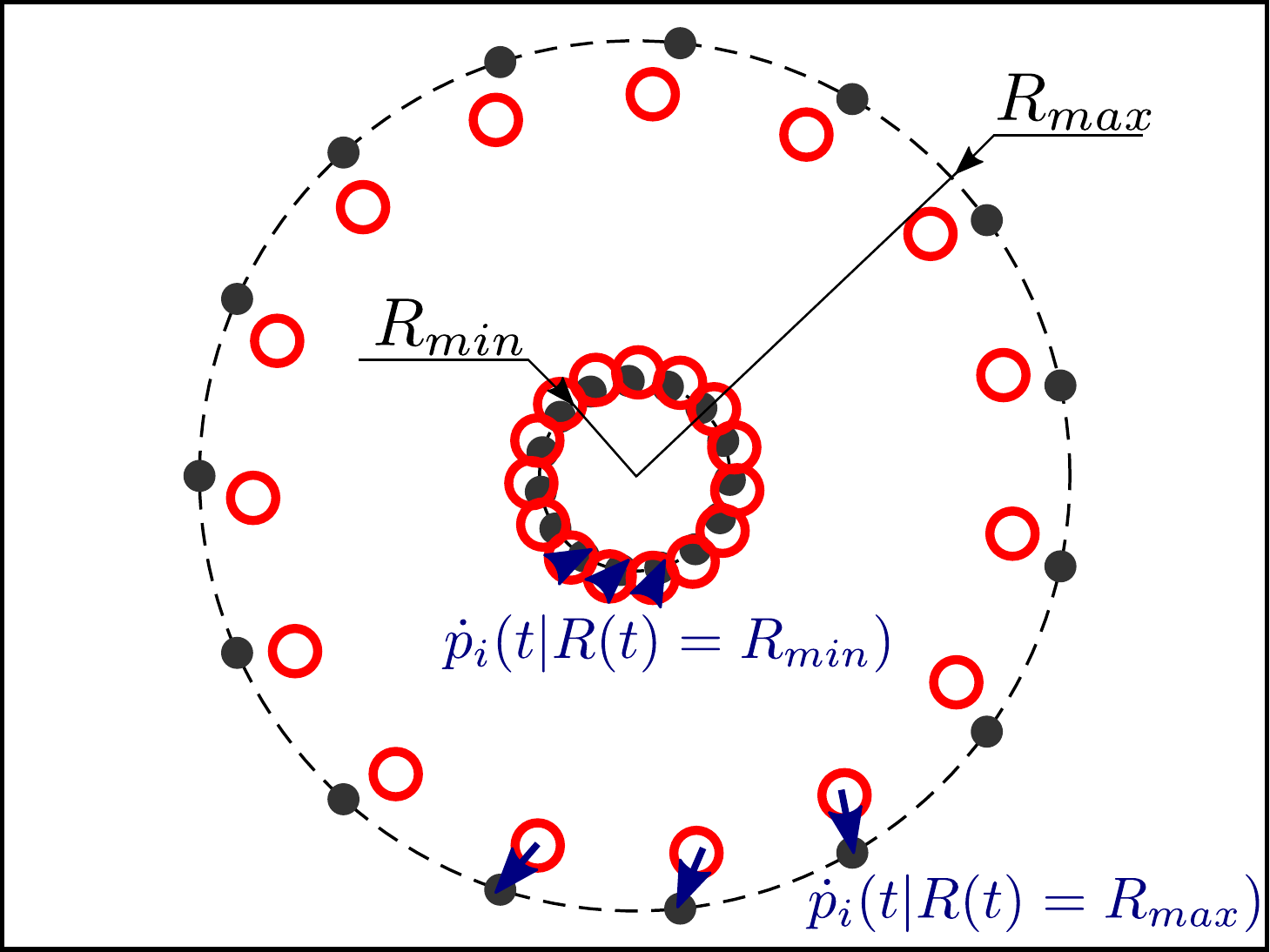}}%
	\hfill%
	\subfloat[\label{subfig:sadnessShape}Sadness: The robots follow points that move slowly along the contour of a small circle with respect to the dimensions of the domain.]{\includegraphics[width=0.32\linewidth]{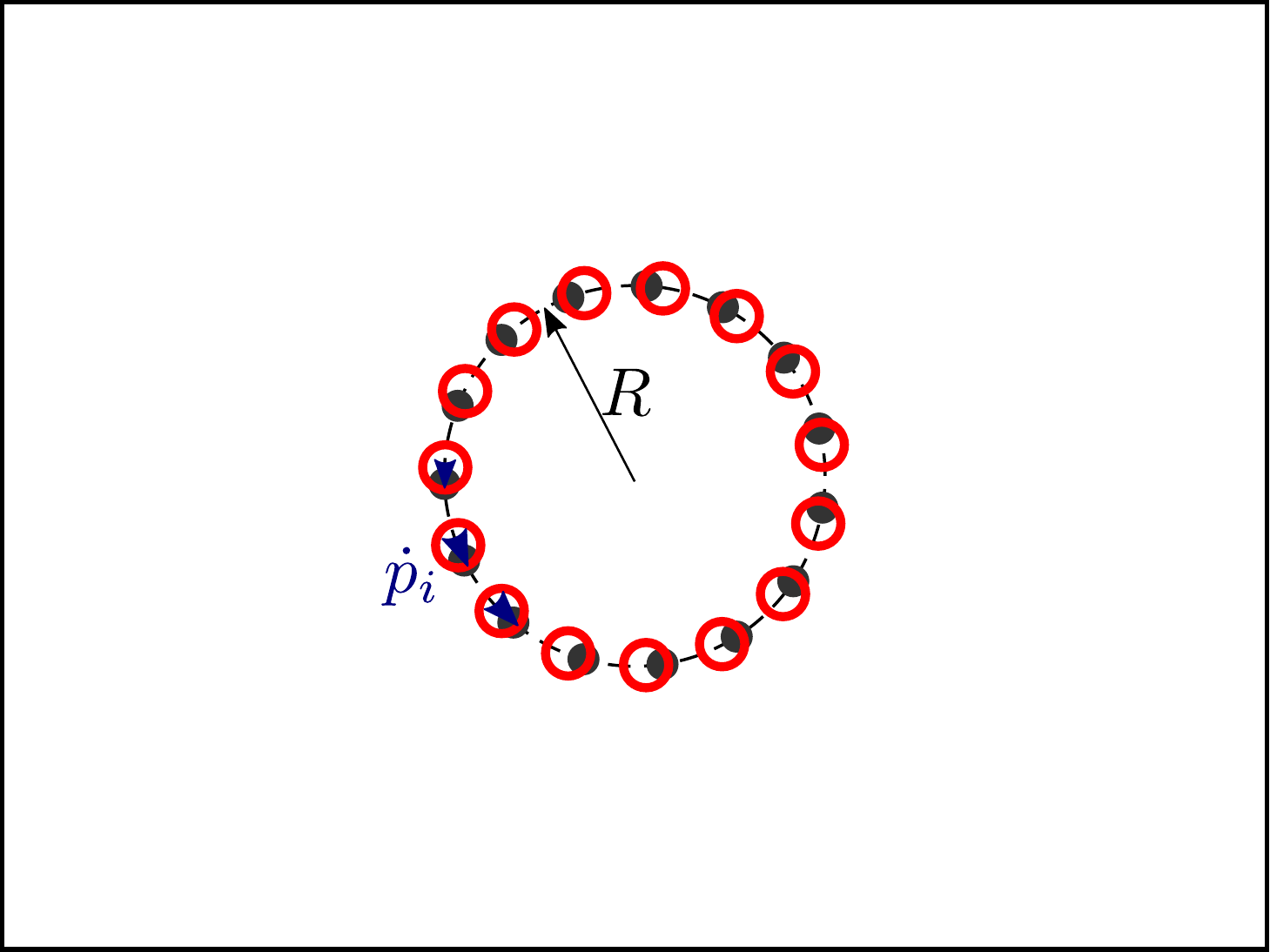}}%
	\caption{Shapes selected for the happiness, surprise and sadness swarm behaviors. Each agent---here depicted as a red circle---follows a point (black circle) that moves along the dashed trajectory. The go-to-go controller that makes each agent follow the corresponding point is illustrated with blue arrows for 3 of the agents.
	}
\end{figure*}

Some confusion was observed in the classification of the behaviors of  fear and disgust, which can be attributed both to the similarity between both emotions in terms of valence and arousal, as well as to the lack of descriptors existent in the literature for the disgust emotion, which complicated the characterization of its associated swarm behavior. Further analysis of the results showed that the robotics background of the participants had no influence on the classification of the behaviors, while the responses of the female participants were more aligned with the hypothesis in comparison to their male counterparts. 

The proposed behaviors were implemented on a team of differential drive robots with the objective of illustrating the feasibility of the proposed behaviors on real robotic platforms. While some differences arose between the simulated and the physical implementation due to the dynamics of the robots, each behavior still displayed its characteristic features. This suggests that the control laws proposed for the different emotions are potentially transferable to other ground robotic systems or even to aerial swarms.

In conclusion, the motion and shape descriptors extracted from social psychology afforded the development of distinct expressive swarm behaviors, identifiable by human observers under one of the fundamental emotions, thus providing a starting point for the design of expressive behaviors for robotic swarms to be used in artistic expositions.

\appendices
\section{Swarm behaviors}
\label{sec:appendix_swarm_behaviors}
In Section \ref{sec:collective_behavior}, a series of swarm behaviors were designed based on the movement and shape attributes associated with the different fundamental emotions. This appendix includes the mathematical expressions of the control laws used to produce the different swarm behaviors. Note that all the control laws included here treat each robot in the swarm as a point that can move omnidirectionally according to single integrator dynamics as in \eqref{eq:single_integrator_dynamics}. The transformation from single integrator dynamics to unicycle dynamics is discussed in detail in Appendix \ref{sec:appendix_diffeomorphism}.

\begin{figure*}[t!]
	\centering%
    \subfloat[\label{subfig:anger} Anger: the Gaussian density makes the robots concentrate around the center of the domain. This choice, along with the selection of a large proportional gain in the diffeomorphism in \eqref{eq:diffeomorphism}, makes the robots stay in each other's vicinity and react to each others movement, producing a jarring movement trace.]{\includegraphics[width=0.32\linewidth]{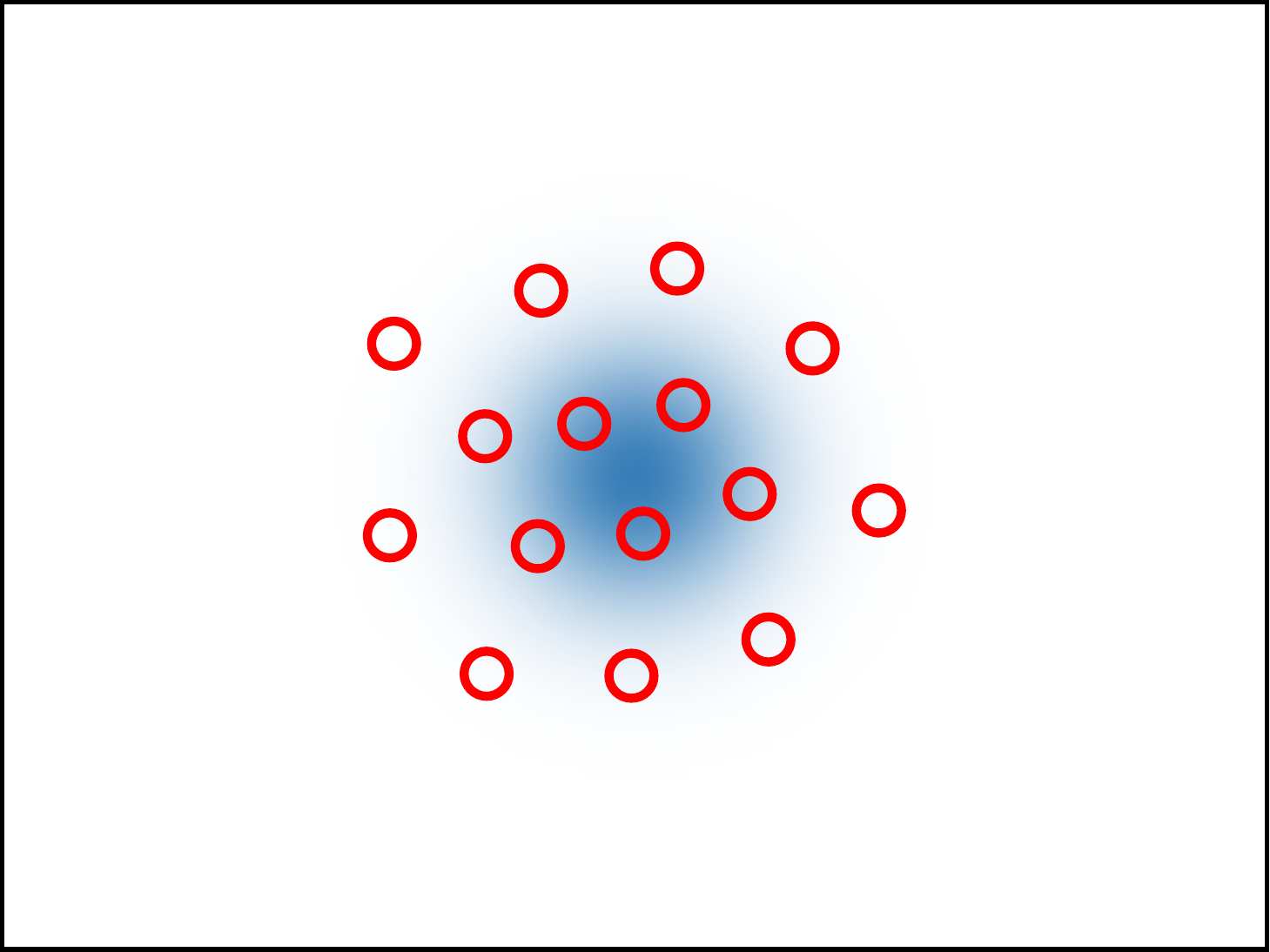}}%
	\hfill%
	\subfloat[\label{subfig:disgust} Disgust: the density function presents high values along the boundaries of the domain. This choice allows the team to spread along the boundary, giving the sensation of animosity between robots.]{\includegraphics[width=0.32\linewidth]{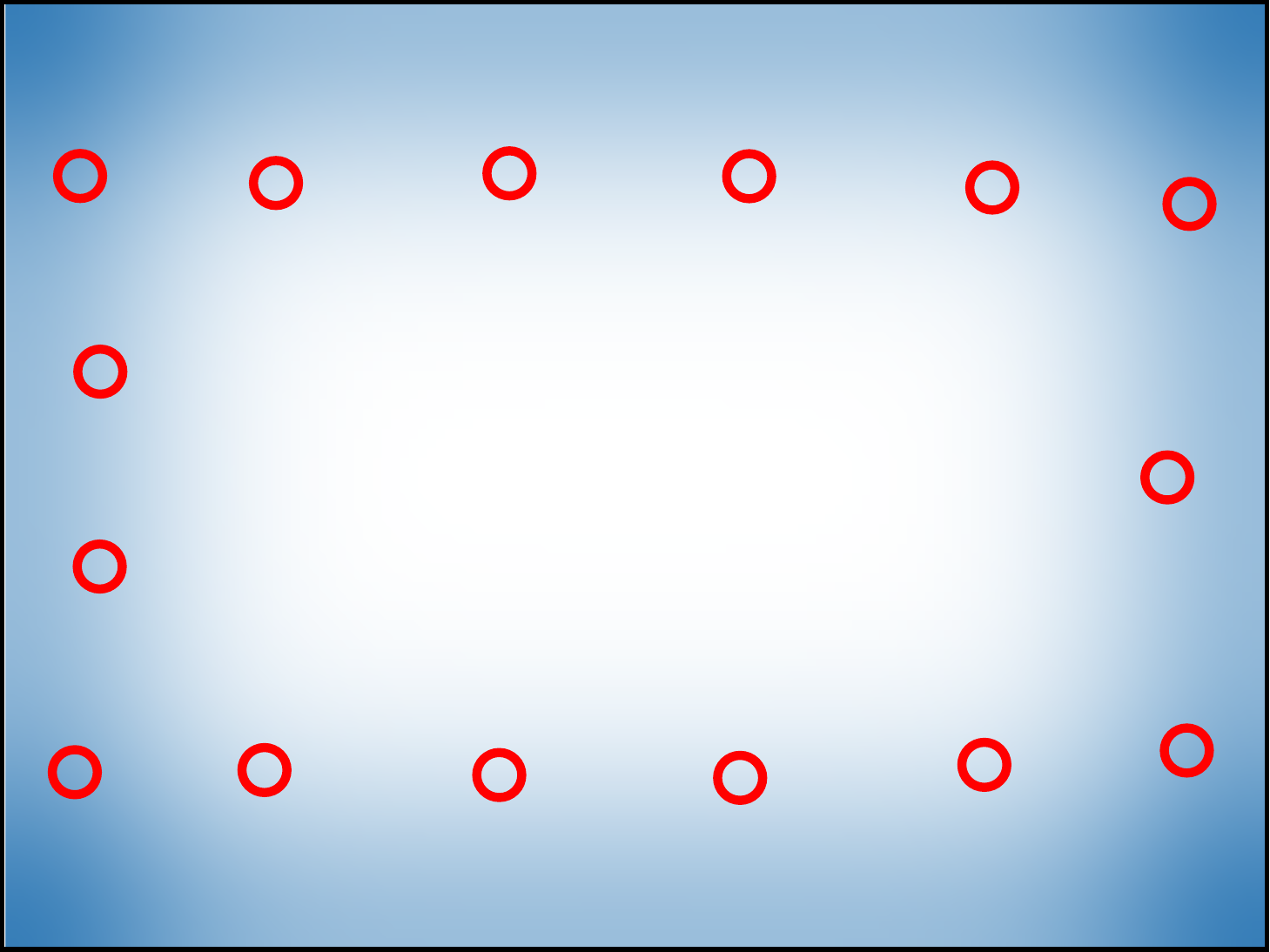}}%
	\hfill%
	\subfloat[\label{subfig:fear} Fear: the density function is chosen to be uniform across the domain. With this choice, the robots scatter evenly over the domain from their initial positions.]{\includegraphics[width=0.32\linewidth]{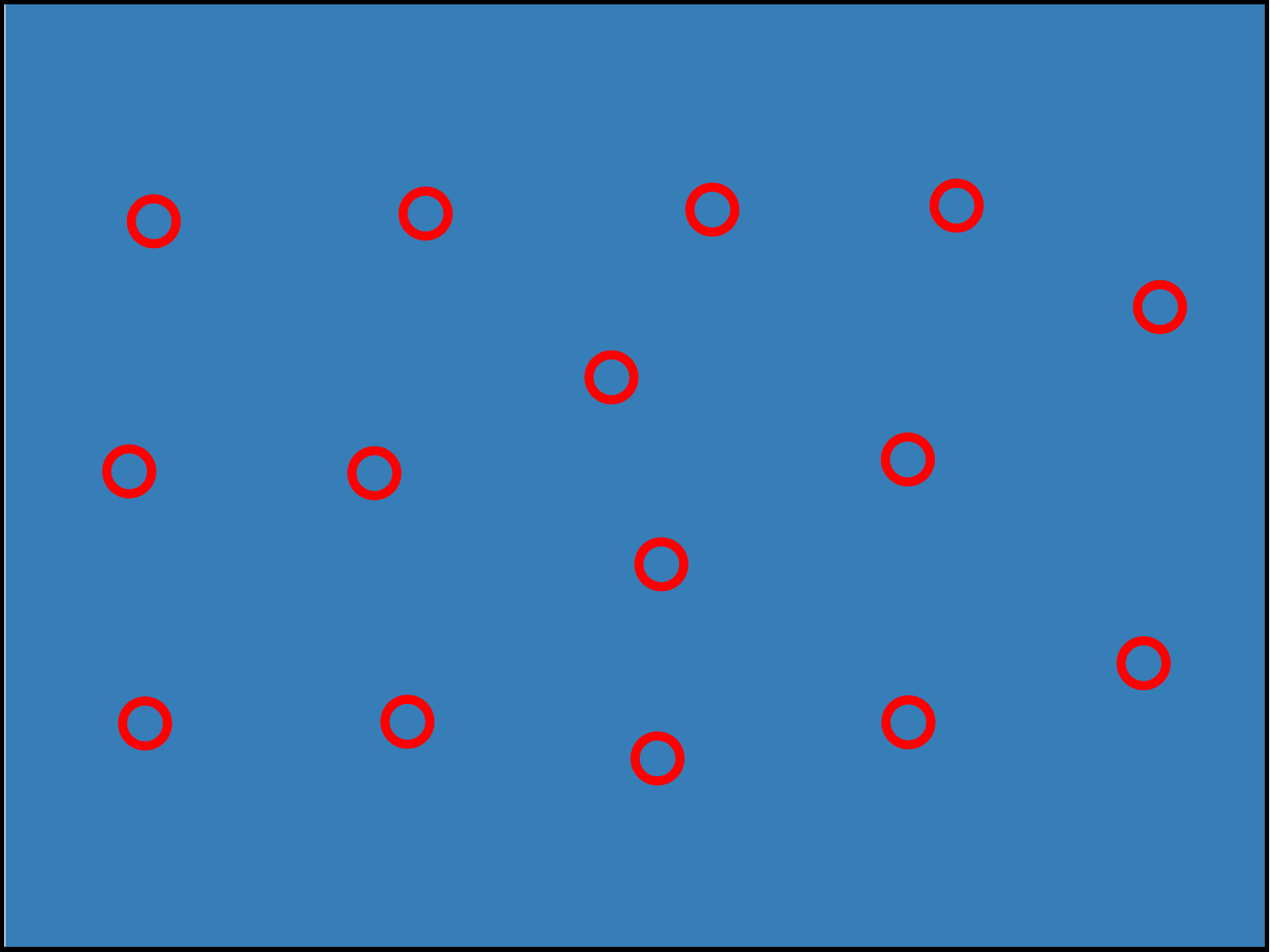}}%
	\caption{Density functions associated to represent the emotions of anger \protect\subref{subfig:anger}, disgust \protect\subref{subfig:disgust} and fear \protect\subref{subfig:fear}. The higher the density (darker color), the higher the concentration of robots will be in that area. The red circles represent the position of the agents once the control law in \eqref{eq:coverage_control} has converged.}
	\label{fig:densities}
\end{figure*}

\subsection{Happiness}
\label{subsec:happiness}
The swarm movement selected for the happiness behavior consists of the robots following the contour of a circle with a superimposed sinusoid. This shape is illustrated in Fig. \ref{subfig:happinessShape} and can be parameterized as
\begin{equation}\label{eq:happiness_shape}
\begin{aligned}
 x_{h}(\theta) & = (R + A\sin(f\theta)) \cos\theta,\\
 y_{h}(\theta) & = (R + A\sin(f\theta)) \sin\theta,
\end{aligned}
\quad \theta\in[0, 2\pi),
\end{equation}
where $R$ is the radius of the main circle and $A$ and $f$ are the amplitude and frequency of the superposed sinusoid, respectively. For the shape in Fig. \ref{subfig:happinessShape}, the frequency of the superimposed sinusoid is $f=6$. 

If we have a swarm of $N$ robots, we can initially position Robot $i$ according to
\begin{equation}\label{eq:happiness_initial_position}
 p_i(0) = [x_h(\theta_i(0)), ~y_h(\theta_i(0))]^T,\quad i=1,\dots, N,
\end{equation}
with
\begin{equation}\label{eq:happiness_initial_angle}
 \theta_i(0) = 2\pi i/N.
\end{equation}
Then the team will depict the desired shape if each robot follows a point evolving along the contour in \eqref{eq:happiness_shape},
\begin{equation}\label{eq:happiness_go_to_goal}
\dot p_i = [x_h(\theta_i(t)), y_h(\theta_i(t))]^T - p_i,
\end{equation}
with $\theta_i$ a function of time $t\in\mathbb{R}_+$,
\begin{equation}\label{eq:theta_between_0_2pi}
\theta_i(t) = \text{atan2}(\sin(t + \theta_i(0)), \cos(t + \theta_i(0))).
\end{equation}

\subsection{Surprise}
In the case of the surprise emotion, each robot follows a point moving along a circle with expanding radius, as in Fig. \ref{subfig:surpriseShape}. Such shape can be parameterized as,
\begin{equation}\label{eq:surprise_shape}
\begin{aligned}
 x_{sur}(\theta, t) & = R(t) \cos\theta,\\
 y_{sur}(\theta, t) & = R(t) \sin\theta,
\end{aligned}
\quad \theta\in[0, 2\pi),
\end{equation}
with
\begin{equation}\label{eq:radius_expanding}
 R(t) = \text{mod}(t, R_{max}-R_{min})+R_{min},\quad t\in\mathbb{R}_+,
\end{equation}
to create a radius that expands from $R_{min}$ to $ R_{max}$.

Analogously to the procedure described in Section \ref{subsec:happiness}, in this case the robots can be initially located at
\begin{equation}\label{eq:surprise_initial_position}
 p_i(0) = [x_{sur}(\theta_i(0), 0), y_{sur}(\theta_i(0), 0)]^T,\quad i=1,\dots, N,
\end{equation}
with $\theta_i(0)$ given by \eqref{eq:happiness_initial_angle}. The controller for each robot is then given by,
\begin{equation}\label{eq:surprise_go_to_goal}
\dot p_i = [x_{sur}(\theta_i(t), 0), y_{sur}(\theta_i(t), 0)]^T - p_i,
\end{equation}
with $\theta_i(t)$ as in \eqref{eq:theta_between_0_2pi}.

\subsection{Sadness}
For the case of the sadness emotion, the robots move along a circle of small dimension as compared to the domain. The strategy is analogous to the ones in \eqref{eq:happiness_go_to_goal} and \eqref{eq:surprise_go_to_goal}, with the parameterization of the contour given by,
\begin{equation}\label{eq:sadness_shape}
\begin{aligned}
x_{sad}(\theta) & = R \cos\theta,\\
y_{sad}(\theta) & = R \sin\theta,
\end{aligned}
\quad \quad \theta\in[0, 2\pi), \quad R>0.
\end{equation}

\subsection{Anger, Fear and Disgust}
For the remaining emotions---anger, disgust and fear---the swarm coordination is based on the \textit{coverage control} strategy, which allows the user to define which areas the robots should concentrate around.

If we denote by $D$ the domain of the robots, the areas where we want to position the robots can be specified by defining a density function, $\phi:D\to[0,\infty)$, that assigns higher values to those areas where we desire the robots to concentrate around. We can make the robots distribute themselves according to this density function by implementing a standard coverage controller such as \cite{Cortes04}, where
\begin{equation}\label{eq:coverage_control}
\dot p_i = \kappa(c_i(p) - p_i), 
\end{equation}
where $p = [p_1^T,\dots,p_N^T]^N$ denotes the aggregate positions of the robots and $\kappa >0$ is a proportional gain. In the controller in \eqref{eq:coverage_control}, $c_i(p)$ denotes the center of mass of the Voronoi cell of Robot $i$,
\begin{equation}\label{eq:coverage_ci}
 c_i(p) = \frac{\int_{V_i(p)}q\phi(q)dq}{\int_{V_i(p)}\phi(q)dq},
\end{equation}
with the Voronoi cell being characterized as,
\begin{equation}\label{eq:Voronoi_cell}
 V_i(p) = \{q\in D ~|~\|q-p_i\|\leq\|q-p_j\|, j\neq i  \}.
\end{equation}
Fig. \ref{fig:densities} shows the densities selected for each of the emotions, where the red circles  represent the positions of the robots in the domain upon convergence, achieved by running the controller in \eqref{eq:coverage_control}.

\section{Individual Robot Control}
\label{sec:appendix_diffeomorphism}

\begin{figure}[t!]
 \centering
 \includegraphics[width=0.6\linewidth]{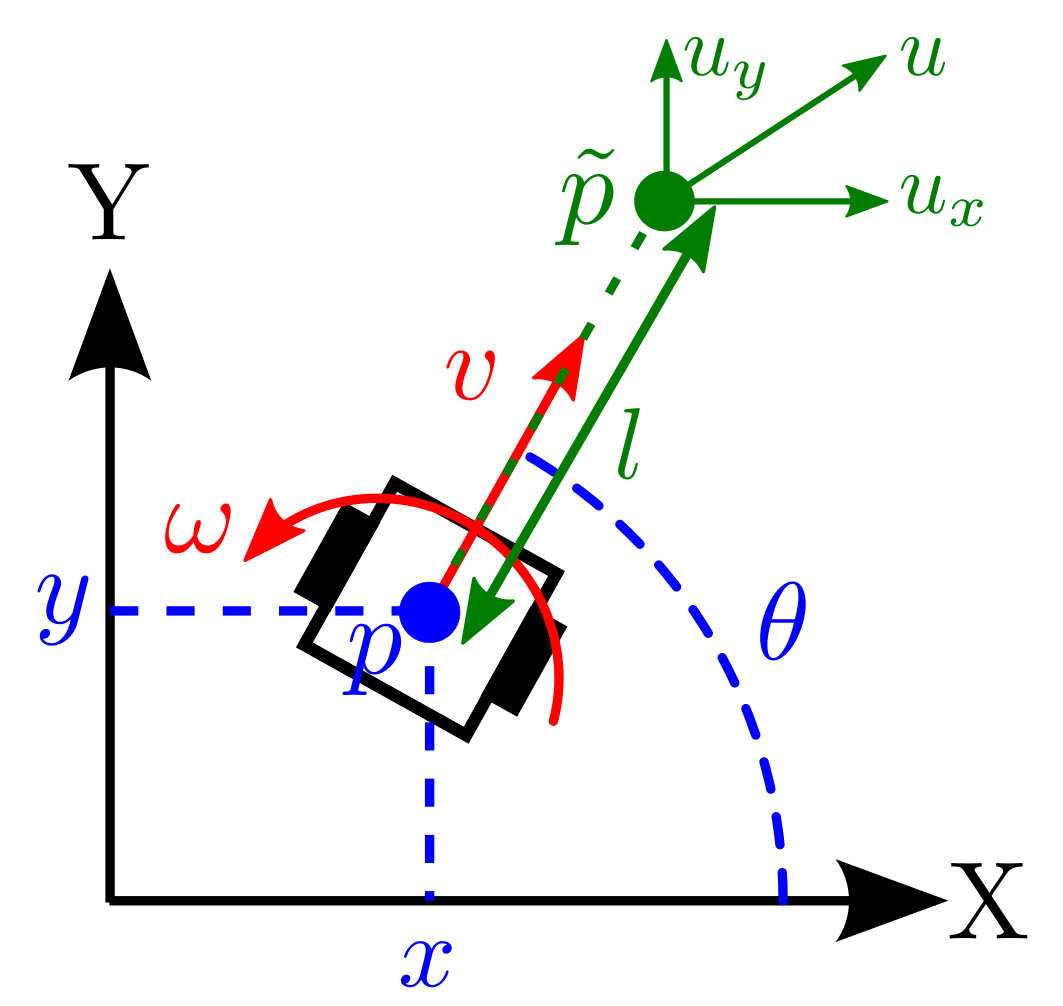}
 \caption{Parameters involved in the near-identity diffeomorphism in \eqref{eq:diffeomorphism}, used to transform the single integrator dynamics in \eqref{eq:single_integrator} into unicycle dynamics \eqref{eq:unicycle_dynamics}, executable by the GRITSBots. The pose of the robot is determined by its position, $p=(x,y)^T$, and its orientation, $\theta$. The single integrator control, $u$, is applied to a point $\tilde p$ located at a distance $l$ in front of the robot. The linear and angular velocities, $v$ and $\omega$, that allow the robot to track $\tilde p$ are obtained applying the near-identity diffeomorphism in \eqref{eq:diffeomorphism}.}
 \label{fig:diffeomorphism}
\end{figure}

The swarm behaviors described in Appendix \ref{sec:appendix_swarm_behaviors} assume that each robot in the swarm can move omnidirectionally according to 
\begin{equation}\label{eq:single_integrator}
 \dot p_i = u_i,
\end{equation}
with $p_i=(x_i, y_i)^T\in\mathbb{R}^2$ the Cartesian position of Robot $i$ in the plane and $u_i=(u_{ix}, u_{iy})^T\in\mathbb{R}^2$ the desired velocity. However, the GRITSBot (Fig. \ref{fig:gritsbot}) has a differential-drive configuration and cannot move omnidirectionally as its motion is constrained in the direction perpendicular to its wheels. Instead, its motion can be expressed as unicycle dynamics,
\begin{align}\label{eq:unicycle_dynamics}
 &\dot x_i = v_i \cos\theta_i,\nonumber\\
 &\dot y_i = v_i \sin\theta_i,\\
 &\dot \theta_i = \omega_i,\nonumber
\end{align}
with $\theta_i$ the orientation of Robot $i$ and $(v_i, \omega_i)^T$ the linear and angular velocities executable by the robot, as shown in Fig. \ref{fig:diffeomorphism}.

In this paper, the single integrator dynamics in \eqref{eq:single_integrator} are converted into unicycle dynamics, as in  \eqref{eq:unicycle_dynamics}, using a near-identity diffeomorphism \cite{Olfati-Saber2002},
\begin{equation}\label{eq:diffeomorphism}
 \begin{pmatrix}
  v_i\\\omega_i
 \end{pmatrix} = K
 \begin{pmatrix}
  \cos \theta_i & \sin \theta_i\\[4pt]
  -\dfrac{\sin\theta_i}{l} & \dfrac{\cos\theta_i}{l}
 \end{pmatrix}
 \begin{pmatrix}
  u_x\\u_y
 \end{pmatrix}, \quad K, l>0.
\end{equation}
A graphical representation of this transformation is included in Fig. \ref{fig:diffeomorphism}: the input $u = (u_x, u_y)^T$ is applied to a point located at a distance of $l$ in front of the robot, $\tilde p$, which can move according to the single integrator dynamics in \eqref{eq:single_integrator}. The effect of this parameter in the movement of the robot is illustrated in Fig. \ref{fig:diffeomorphism_trajectory}. The parameter $K$ acts as a proportional gain.

\bibliographystyle{IEEEtran}
\bibliography{biblio}

\begin{thebibliography}{10}
\providecommand{\url}[1]{#1}
\csname url@rmstyle\endcsname
\providecommand{\newblock}{\relax}
\providecommand{\bibinfo}[2]{#2}
\providecommand\BIBentrySTDinterwordspacing{\spaceskip=0pt\relax}
\providecommand\BIBentryALTinterwordstretchfactor{4}
\providecommand\BIBentryALTinterwordspacing{\spaceskip=\fontdimen2\font plus
\BIBentryALTinterwordstretchfactor\fontdimen3\font minus
  \fontdimen4\font\relax}
\providecommand\BIBforeignlanguage[2]{{%
\expandafter\ifx\csname l@#1\endcsname\relax
\typeout{** WARNING: IEEEtran.bst: No hyphenation pattern has been}%
\typeout{** loaded for the language `#1'. Using the pattern for}%
\typeout{** the default language instead.}%
\else
\language=\csname l@#1\endcsname
\fi
#2}}

\bibitem{Brown2013}
L.~Brown, R.~Kerwin, and A.~M. Howard, ``Applying behavioral strategies for
  student engagement using a robotic educational agent,'' in \emph{2013 IEEE
  International Conference on Systems, Man, and Cybernetics}, Oct. 2013, pp.
  4360--4365.

\bibitem{Belpaeme2013}
T.~Belpaeme, P.~Baxter, R.~Read, R.~Wood, H.~Cuay\'{a}huitl, B.~Kiefer,
  S.~Racioppa, I.~Kruijff-Korbayov\'{a}, G.~Athanasopoulos, V.~Enescu,
  R.~Looije, M.~Neerincx, Y.~Demiris, R.~Ros-Espinoza, A.~Beck,
  L.~Ca\~{n}amero, A.~Hiolle, M.~Lewis, I.~Baroni, M.~Nalin, P.~Cosi, G.~Paci,
  F.~Tesser, G.~Sommavilla, and R.~Humbert, ``Multimodal child-robot
  interaction: Building social bonds,'' \emph{J. Hum.-Robot Interact.}, vol.~1,
  no.~2, pp. 33--53, Jan. 2013.

\bibitem{Hoffman2013}
G.~Hoffman, ``Dumb robots, smart phones: A case study of music listening
  companionship,'' in \emph{2012 IEEE RO-MAN: The 21st IEEE International
  Symposium on Robot and Human Interactive Communication}, Sept 2012, pp.
  358--363.

\bibitem{Cabibihan2013}
J.-J. Cabibihan, H.~Javed, M.~Ang, and S.~M. Aljunied, ``Why robots? a survey
  on the roles and benefits of social robots in the therapy of children with
  autism,'' \emph{International Journal of Social Robotics}, vol.~5, no.~4, pp.
  593--618, Nov 2013.

\bibitem{Kozima2009}
H.~Kozima, M.~P. Michalowski, and C.~Nakagawa, ``Keepon,'' \emph{International
  Journal of Social Robotics}, vol.~1, no.~1, pp. 3--18, Jan. 2009.

\bibitem{Breazeal2003}
C.~Breazeal, ``Toward sociable robots,'' \emph{Robotics and Autonomous
  Systems}, vol.~42, pp. 167--175, 2003.

\bibitem{Hoffman2010}
G.~Hoffman and G.~Weinberg, ``Gesture-based human-robot jazz improvisation,''
  in \emph{2010 IEEE International Conference on Robotics and Automation}, May
  2010, pp. 582--587.

\bibitem{Bi2018}
T.~Bi, P.~Fankhauser, D.~Bellicoso, and M.~Hutter, ``Real-time dance generation
  to music for a legged robot,'' in \emph{2018 IEEE/RSJ International
  Conference on Intelligent Robots and Systems (IROS)}, Madrid, Oct. 2018, pp.
  1038--1044.

\bibitem{LaViers2014}
A.~LaViers, L.~Teague, and M.~Egerstedt, \emph{Style-Based Robotic Motion in
  Contemporary Dance Performance}.\hskip 1em plus 0.5em minus 0.4em\relax Cham:
  Springer International Publishing, 2014, pp. 205--229.

\bibitem{Nakazawa2002}
A.~Nakazawa, S.~Nakaoka, K.~Ikeuchi, and K.~Yokoi, ``Imitating human dance
  motions through motion structure analysis,'' in \emph{IEEE/RSJ International
  Conference on Intelligent Robots and Systems}, vol.~3, Oct. 2002, pp.
  2539--2544.

\bibitem{Shinozaki2008}
K.~Shinozaki, A.~Iwatani, and R.~Nakatsu, ``Construction and evaluation of a
  robot dance system,'' in \emph{RO-MAN 2008 - The 17th IEEE International
  Symposium on Robot and Human Interactive Communication}, Aug. 2008, pp.
  366--370.

\bibitem{Lee2014}
D.~Lee, S.~Park, M.~Hahn, and N.~Lee, ``Robot actors and authoring tools for
  live performance system,'' in \emph{2014 International Conference on
  Information Science Applications (ICISA)}, May 2014, pp. 1--3.

\bibitem{Perkowski2005}
M.~Perkowski, T.~Sasao, J.~H. Kim, M.~Lukac, J.~Allen, and S.~Gebauer, ``{Hahoe
  KAIST Robot Theatre}: learning rules of interactive robot behavior as a
  multiple-valued logic synthesis problem,'' in \emph{35th International
  Symposium on Multiple-Valued Logic (ISMVL'05)}, May 2005, pp. 236--248.

\bibitem{Sunardi2018}
M.~Sunardi and M.~Perkowski, ``Music to motion: Using music information to
  create expressive robot motion,'' \emph{International Journal of Social
  Robotics}, vol.~10, no.~1, pp. 43--63, Jan 2018.

\bibitem{Ackerman2014}
E.~Ackerman, ``Flying {L}ampshade{B}ots {C}ome {A}live in {C}irque du
  {S}oleil,'' \emph{IEEE Spectrum}, 2014.

\bibitem{Dean2008}
M.~Dean, R.~D'Andrea, and M.~Donovan, \emph{Robotic Chair}.\hskip 1em plus
  0.5em minus 0.4em\relax Vancouver, B.C.: Contemporary Art Gallery, 2008.

\bibitem{Dunstan2016}
B.~J. Dunstan, D.~Silvera-Tawil, J.~T. K.~V. Koh, and M.~Velonaki, ``Cultural
  robotics: Robots as participants and creators of culture,'' in \emph{Cultural
  Robotics}, J.~T. Koh, B.~J. Dunstan, D.~Silvera-Tawil, and M.~Velonaki,
  Eds.\hskip 1em plus 0.5em minus 0.4em\relax Cham: Springer International
  Publishing, 2016, pp. 3--13.

\bibitem{Vlachos2018}
E.~{Vlachos}, E.~{Jochum}, and L.~{Demers}, ``Heat: The harmony exoskeleton
  self - assessment test,'' in \emph{2018 27th IEEE International Symposium on
  Robot and Human Interactive Communication (RO-MAN)}, Aug 2018, pp. 577--582.

\bibitem{Camurri2004}
A.~Camurri, B.~Mazzarino, M.~Ricchetti, R.~Timmers, and G.~Volpe, ``Multimodal
  analysis of expressive gesture in music and dance performances,'' in
  \emph{Gesture-Based Communication in Human-Computer Interaction}, A.~Camurri
  and G.~Volpe, Eds.\hskip 1em plus 0.5em minus 0.4em\relax Berlin, Heidelberg:
  Springer Berlin Heidelberg, 2004, pp. 20--39.

\bibitem{Juslin2005}
P.~Juslin, \emph{From mimesis to catharsis: expression, perception, and
  induction of emotion in music}.\hskip 1em plus 0.5em minus 0.4em\relax
  Oxford: Oxford University Press, 2005.

\bibitem{Or2009}
J.~Or, ``Towards the development of emotional dancing humanoid robots,''
  \emph{International Journal of Social Robotics}, vol.~1, no.~4, p. 367, Oct.
  2009.

\bibitem{Perkowski2013}
M.~Perkowski, A.~Bhutada, M.~Lukac, and M.~Sunardi, ``On synthesis and
  verification from event diagrams in a robot theatre application,'' in
  \emph{2013 IEEE 43rd International Symposium on Multiple-Valued Logic}, May
  2013, pp. 77--83.

\bibitem{Bretan2015}
M.~Bretan, G.~Hoffman, and G.~Weinberg, ``Emotionally expressive dynamic
  physical behaviors in robots,'' \emph{International Journal of Human-Computer
  Studies}, vol.~78, pp. 1 -- 16, 2015.

\bibitem{Hoffman2008}
G.~Hoffman, R.~Kubat, and C.~Breazeal, ``A hybrid control system for
  puppeteering a live robotic stage actor,'' in \emph{RO-MAN 2008 - The 17th
  IEEE International Symposium on Robot and Human Interactive Communication},
  Aug 2008, pp. 354--359.

\bibitem{Schoellig2014}
A.~P. Schoellig, H.~Siegel, F.~Augugliaro, and R.~D'Andrea, \emph{So You Think
  You Can Dance? Rhythmic Flight Performances with Quadrocopters}.\hskip 1em
  plus 0.5em minus 0.4em\relax Cham: Springer International Publishing, 2014,
  pp. 73--105.

\bibitem{Dietz2017}
G.~Dietz, J.~L. E, P.~Washington, L.~H. Kim, and S.~Follmer, ``Human perception
  of swarm robot motion,'' in \emph{Proceedings of the 2017 CHI Conference
  Extended Abstracts on Human Factors in Computing Systems}, Denver, Colorado,
  2017, pp. 2520--2527.

\bibitem{Levillain2018}
F.~{Levillain}, D.~{St-Onge}, E.~{Zibetti}, and G.~{Beltrame}, ``More than the
  sum of its parts: Assessing the coherence and expressivity of a robotic
  swarm,'' in \emph{2018 IEEE International Symposium on Robot and Human
  Interactive Communication (RO-MAN)}, Aug 2018, pp. 583--588.

\bibitem{StOnge2019}
D.~St-Onge, F.~Levillain, Z.~Elisabetta, and G.~Beltrame, ``Collective
  expression: how robotic swarms convey information with group motion,''
  \emph{Paladyn, Journal of Behavioral Robotics}, vol.~10, pp. 418--435, 12
  2019.

\bibitem{Goodrich2007HRIsurvey}
M.~A. Goodrich and A.~C. Schultz, ``Human-robot interaction: A survey,''
  \emph{Found. Trends Hum.-Comput. Interact.}, vol.~1, no.~3, pp. 203--275,
  Jan. 2007.

\bibitem{Kolling2016}
A.~Kolling, P.~Walker, N.~Chakraborty, K.~Sycara, and M.~Lewis, ``Human
  interaction with robot swarms: A survey,'' \emph{IEEE Transactions on
  Human-Machine Systems}, vol.~46, no.~1, pp. 9--26, Feb. 2016.

\bibitem{Sheridan2016}
T.~B. Sheridan, ``Human–robot interaction: Status and challenges,''
  \emph{Human Factors}, vol.~58, no.~4, pp. 525--532, 2016.

\bibitem{Alonso-Mora2014}
J.~Alonso-Mora, R.~Siegwart, and P.~Beardsley, ``{H}uman-{R}obot swarm
  interaction for entertainment: From animation display to gesture based
  control,'' in \emph{Proceedings of the 2014 ACM/IEEE International Conference
  on Human-robot Interaction}, ser. HRI '14.\hskip 1em plus 0.5em minus
  0.4em\relax New York, NY, USA: ACM, 2014, pp. 98--98.

\bibitem{Collier1996}
G.~L. Collier, ``Affective synesthesia: Extracting emotion space from simple
  perceptual stimuli,'' \emph{Motivation and Emotion}, vol.~20, no.~1, pp.
  1--32, 1996.

\bibitem{DeRooij2013}
A.~de~Rooij, J.~Broekens, and M.~H. Lamers, ``Abstract expressions of affect,''
  \emph{International Journal of Synthetic Emotions}, vol.~4, no.~1, pp. 1--31,
  2013.

\bibitem{Lee2007}
J.-H. Lee, J.-Y. Park, and T.-J. Nam, \emph{Emotional Interaction Through
  Physical Movement}.\hskip 1em plus 0.5em minus 0.4em\relax Berlin,
  Heidelberg: Springer Berlin Heidelberg, 2007, pp. 401--410.

\bibitem{Pollick2001}
F.~E. Pollick, H.~M. Paterson, A.~Bruderlin, and A.~J. Sanford, ``Perceiving
  affect from arm movement,'' \emph{Cognition}, vol.~82, no.~2, pp. B51 -- B61,
  2001.

\bibitem{Rime1985}
B.~Rim{\'e}, B.~Boulanger, P.~Laubin, M.~Richir, and K.~Stroobants, ``The
  perception of interpersonal emotions originated by patterns of movement,''
  \emph{Motivation and Emotion}, vol.~9, no.~3, pp. 241--260, 1985.

\bibitem{Aronoff2006}
J.~Aronoff, ``How we recognize angry and happy emotion in people, places, and
  things,'' \emph{Cross-Cultural Research}, vol.~40, no.~1, pp. 83--105, 2006.

\bibitem{Ekman1993}
P.~Ekman, ``Facial expression and emotion,'' \emph{American Psychologist},
  vol.~48, no.~4, pp. 384--392, 1993.

\bibitem{Izard2009}
C.~E. Izard, ``Emotion theory and research: Highlights, unanswered questions,
  and emerging issues,'' \emph{Annual review of psychology}, vol.~60, pp.
  1--25, 2009.

\bibitem{Izard1977}
------, \emph{Human Emotions}, ser. Emotions, Personalityand
  Psychotherapy.\hskip 1em plus 0.5em minus 0.4em\relax New York, NY, USA:
  Springer US, 1977.

\bibitem{Plutchik2001}
R.~Plutchik, ``The nature of emotions: Human emotions have deep evolutionary
  roots, a fact that may explain their complexity and provide tools for
  clinical practice,'' \emph{American Scientist}, vol.~89, no.~4, pp. 344--350,
  2001.

\bibitem{Pickem15}
D.~Pickem, M.~Lee, and M.~Egerstedt, ``The {GRITSB}ot in its natural habitat --
  a multi-robot testbed,'' in \emph{2015 IEEE International Conference on
  Robotics and Automation (ICRA)}, Seattle, WA, May 2015, pp. 4062--4067.

\bibitem{Laban1947}
R.~Laban and F.~Lawrence, \emph{Effort}.\hskip 1em plus 0.5em minus 0.4em\relax
  London: Macdonald \& Evans Ltd, 1947.

\bibitem{Barakova2010}
E.~I. Barakova and T.~Lourens, ``Expressing and interpreting emotional
  movements in social games with robots,'' \emph{Personal and Ubiquitous
  Computing}, vol.~14, no.~5, pp. 457--467, July 2010.

\bibitem{Knight2014}
H.~{Knight} and R.~{Simmons}, ``Expressive motion with x, y and theta: Laban
  effort features for mobile robots,'' in \emph{The 23rd IEEE International
  Symposium on Robot and Human Interactive Communication}, Aug 2014, pp.
  267--273.

\bibitem{Lourens2010}
T.~Lourens, R.~van Berkel, and E.~Barakova, ``Communicating emotions and mental
  states to robots in a real time parallel framework using laban movement
  analysis,'' \emph{Robotics and Autonomous Systems}, vol.~58, no.~12, pp. 1256
  -- 1265, 2010, intelligent Robotics and Neuroscience.

\bibitem{Masuda2010}
M.~Masuda, S.~Kato, and H.~Itoh, ``A laban-based approach to emotional motion
  rendering for human-robot interaction,'' in \emph{Entertainment Computing -
  ICEC 2010}, H.~S. Yang, R.~Malaka, J.~Hoshino, and J.~H. Han, Eds.\hskip 1em
  plus 0.5em minus 0.4em\relax Berlin, Heidelberg: Springer Berlin Heidelberg,
  2010, pp. 372--380.

\bibitem{Frijda1986}
N.~Frijda, \emph{The Emotions}, ser. Studies in Emotion and Social
  Interaction.\hskip 1em plus 0.5em minus 0.4em\relax Cambridge University
  Press, 1986.

\bibitem{Russell1980}
J.~A. Russell, ``A circumplex model of affect,'' \emph{Journal of Personality
  and Social Psychology}, vol.~39, no.~6, pp. 1161--1178, 1980.

\bibitem{Justh2003}
E.~W. {Justh} and P.~S. {Krishnaprasad}, ``Steering laws and continuum models
  for planar formations,'' in \emph{42nd IEEE International Conference on
  Decision and Control}, vol.~4, Dec 2003, pp. 3609--3614 vol.4.

\bibitem{Marshall2004}
J.~A. {Marshall}, M.~E. {Broucke}, and B.~A. {Francis}, ``Formations of
  vehicles in cyclic pursuit,'' \emph{IEEE Transactions on Automatic Control},
  vol.~49, no.~11, pp. 1963--1974, Nov 2004.

\bibitem{Ramirez2009}
J.~L. {Ramirez}, M.~{Pavone}, E.~{Frazzoli}, and D.~W. {Miller}, ``Distributed
  control of spacecraft formation via cyclic pursuit: Theory and experiments,''
  in \emph{2009 American Control Conference}, June 2009, pp. 4811--4817.

\bibitem{Cortes04}
J.~Cortes, S.~Martinez, T.~Karatas, and F.~Bullo, ``Coverage control for mobile
  sensing networks,'' \emph{IEEE Transactions on Robotics and Automation},
  vol.~20, no.~2, pp. 243--255, Apr. 2004.

\bibitem{DiazMercado2015}
Y.~{Diaz-Mercado}, S.~G. {Lee}, and M.~{Egerstedt}, ``Distributed dynamic
  density coverage for human-swarm interactions,'' in \emph{2015 American
  Control Conference (ACC)}, July 2015, pp. 353--358.

\bibitem{Olfati-Saber2002}
R.~Olfati-Saber, ``Near-identity diffeomorphisms and exponential
  $\epsilon$-tracking and $\epsilon$-stabilization of first-order nonholonomic
  se(2) vehicles,'' in \emph{Proceedings of the 2002 American Control
  Conference}, vol.~6, May 2002, pp. 4690--4695.

\bibitem{Ross1938}
R.~T. Ross, ``A statistic for circular series,'' \emph{Journal of Educational
  Psychology}, vol.~29, no.~5, pp. 384--389, 1938.

\bibitem{Wilson2020}
S.~{Wilson}, P.~{Glotfelter}, L.~{Wang}, S.~{Mayya}, G.~{Notomista}, M.~{Mote},
  and M.~{Egerstedt}, ``The robotarium: Globally impactful opportunities,
  challenges, and lessons learned in remote-access, distributed control of
  multirobot systems,'' \emph{IEEE Control Systems Magazine}, vol.~40, no.~1,
  pp. 26--44, Feb 2020.

\end{thebibliography}

\end{document}